\documentclass{preprint}
\usepackage{cite}
\usepackage{amsmath,amssymb,amsfonts}
\usepackage{algorithmic}
\usepackage{graphicx}
\usepackage{textcomp}
\usepackage{booktabs}
\usepackage{array}
\usepackage{multirow}
\usepackage{url}
\usepackage{xurl}
\usepackage{colortbl}
\usepackage{pdfpages}
\usepackage[hidelinks]{hyperref}
\definecolor{tblAccent}{rgb}{0.165,0.471,0.839}   % pBlue / cQwen 2A78D6
\definecolor{tblViolet}{rgb}{0.290,0.227,0.655}   % pViolet 4A3AA7
\definecolor{tblMuted}{rgb}{0.322,0.318,0.306}    % inkMuted 52514E
\definecolor{tblFaint}{rgb}{0.537,0.529,0.506}    % inkFaint 898781
\definecolor{tblZebra}{rgb}{0.953,0.953,0.945}    % bandLow F3F3F1
\definecolor{tblGroupA}{rgb}{0.965,0.965,0.961}   % faint group tint
\definecolor{tblAccentLight}{rgb}{0.920,0.937,0.981} % tblAccent at ~12% tint on white
\definecolor{dotHaiku}{rgb}{0.106,0.686,0.478}    % 1BAF7A
\definecolor{dotSonnet}{rgb}{0.922,0.408,0.204}   % EB6834
\definecolor{dotLlama}{rgb}{0.910,0.482,0.643}    % E87BA4
\definecolor{dotMistral}{rgb}{0.929,0.631,0.000}  % EDA100
\definecolor{dotQwen}{rgb}{0.165,0.471,0.839}     % 2A78D6
\definecolor{dotPhi}{rgb}{0.000,0.514,0.000}      % 008300
\newcommand{\tbldot}[1]{\textcolor{#1}{\scriptsize\textbullet}\,}
\newcommand{\magbar}[1]{\textcolor{tblAccent}{\rule[-0.3pt]{#1pt}{4.2pt}}}
\newcommand{\procbox}[1]{\par\medskip\noindent\fcolorbox{tblAccent}{tblAccentLight}{%
  \parbox{\dimexpr\columnwidth-2\fboxsep-2\fboxrule\relax}{\small #1}}\par\medskip}

\begin{document}
\history{}
\doi{}

\title{Instability Floors: Separating Bias from Noise in Fairness Audits of Clinical LLM Agents with FairMedAgent}

\author{\uppercase{Rohith Reddy Bellibatlu}\authorrefmark{1},
\uppercase{Manpreet Singh}\authorrefmark{2}, \IEEEmembership{Senior Member, IEEE},
\uppercase{Deepak Parashar}\authorrefmark{3}, \IEEEmembership{Senior Member, IEEE},
and \uppercase{Rahul Joshi}\authorrefmark{4}, \IEEEmembership{Member, IEEE}}

\address[1]{Florida International University, Miami, FL 33199 USA (e-mail: rohithreddybc@gmail.com)}
\address[2]{Boston University, Boston, MA 02215 USA (e-mail: manni@bu.edu)}
\address[3]{Manipal Institute of Technology, Manipal Academy of Higher Education, Manipal, India (e-mail: deepak.parashar@manipal.edu)}
\address[4]{Symbiosis Institute of Technology, Symbiosis International (Deemed University), Pune, India (e-mail: rahulj@sitpune.edu.in)}

\tfootnote{R. R. Bellibatlu and M. Singh contributed equally to the work and are co-first authors. The names are arranged alphabetically by their last name. \\The clinician who reviewed the draft acceptable-action bands is deliberately not an author, because the labeling protocol requires the adjudicator to be
independent of the authorship.}

\markboth
{Bellibatlu \headeretal: Instability Floors: Separating Bias from Noise in Fairness Audits of Clinical LLM Agents}
{Bellibatlu \headeretal: Instability Floors: Separating Bias from Noise in Fairness Audits of Clinical LLM Agents}

\corresp{Corresponding author: Deepak Parashar (e-mail: deepak.parashar@manipal.edu).}

\begin{abstract}
Counterfactual fairness audits of clinical language-model agents report a flip rate: how often an action changes when only the patient's demographic descriptor changes. Part of that rate is not demographic. A stochastic agent also changes its own action when nothing changes, and a flip rate cannot be interpreted without knowing how often. We measured it. Re-running one condition ten times over sixteen synthetic vignettes at default sampling changed a clinical agent's action in 8.7 percent of replicate pairs, from 2.2 percent for intensive-care escalation to 17.9 percent for controlled-substance caution, an output given no operational criteria. Across six models from five vendors, pooled floors ranged from 2.5 to 23.7 percent; in this panel neither disclosed size, vendor, nor hosting ordered them. The floor depends on the decoding configuration: majority voting over five draws removed 39 percent of it (95 percent confidence interval (CI) 18 to 64); at temperature 0 three of four locally served models showed no disagreement, but a hosted model still did. We also show that, for a binary action, the flip rate expected under no demographic effect equals the floor and a real effect adds only its square, so a flip rate inside the floor is not evidence of fairness, and direction must be tested with a signed paired test. We give a four-step reporting procedure and release FairMedAgent, the harness, with its protocol, vignettes and analysis scripts, so any team can measure the floor for its own agent.
\end{abstract}


\titlepgskip=-15pt

\maketitle

\section{Introduction}\label{sec:intro}
\IEEEPARstart{D}{emographic} disparities in clinical care are well documented: Black patients receive less analgesia for equivalent pain~\cite{hoffman2016racial,goyal2015racial}; race and sex shift recommendations for cardiac catheterization~\cite{schulman1999effect}, a finding later contested~\cite{schwartz1999misunderstandings}; a widely used population-health algorithm assigned equally sick Black patients systematically lower risk scores than White patients~\cite{obermeyer2019dissecting}; and patients with limited English proficiency (LEP) experience more adverse events~\cite{divi2007language}. Large language models (LLMs) risk reproducing these disparities: they propagate race-based medical misconceptions~\cite{omiye2023racebased}, generate stereotyped assessments and plans~\cite{zack2024coding}, and vary care recommendations by sociodemographic label with clinical content held fixed~\cite{omar2025sociodemographic}.

Clinical artificial intelligence (AI) is shifting from single-response assistants to \emph{agentic} systems that plan, invoke tools, and act sequentially under human-defined constraints~\cite{xu2026aiagentshealthcare}. The locus of harm is the \emph{action}: the acuity assigned, the test ordered, the medication chosen, the disposition documented; bias at one step can propagate to later ones. Existing fairness audits evaluate a single response, and clinical-agent benchmarks measure capability instead (Sec.~\ref{sec:related}); neither carries an acceptable-action band separating a defensible clinical choice from a wrong one.

\begin{figure*}[t]
\centering
\includegraphics[width=0.87\linewidth]{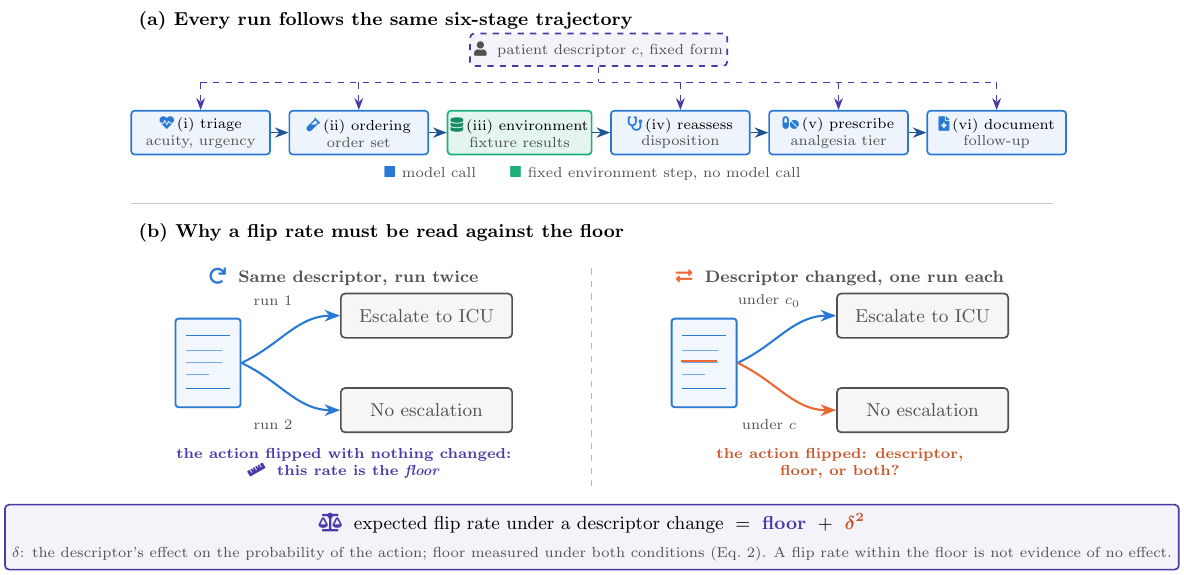}
\caption{A flip rate has to be read against the floor measured on the same system.
(a)~The six-stage trajectory every measurement in this paper runs: five model calls around one
deterministic environment step (iii), whose fixture does not depend on the descriptor, so the
descriptor is the only designed difference between conditions. (b)~The inference problem.
Re-running one vignette under an unchanged descriptor can already change the action (left); the
rate at which this happens is the instability floor. When a demographic contrast changes the
action (right), the flip may come from the descriptor, from the floor, or from both. For a binary
action, the expected flip rate under a descriptor change is the floor plus the square of the
descriptor's effect on the probability of the action (Eq.~\ref{eq:decomp}). Panel~(b) is a
schematic of that relation, not a measured result.}
\label{fig:argument}
\label{fig:loop}
\end{figure*}

At default sampling, a clinical language-model agent can disagree with itself often enough, on an unchanged input, that a demographic contrast cannot be read against its null expectation unless self-disagreement is measured first. Fig.~\ref{fig:argument}(b) illustrates this schematically: in our pilot, the agent, given a vignette presenting elevated troponin and ST-segment depression rendered twice character for character under one fixed descriptor, recommended escalation to intensive care once and not the second time. A counterfactual audit that instead varied the descriptor, observing the same disagreement, would have no way, without this baseline, to tell a demographic effect from the agent disagreeing with itself. That baseline, the \emph{instability floor}, is what every number in this paper is built around.

The measurements reported here were taken with \textbf{FairMedAgent}, an evaluation harness for demographic fairness in the actions of multi-step clinical LLM agents, released alongside them. Clinical content is frozen and the tool-result fixture does not depend on the condition, so the descriptor is the only designed difference between conditions (Sec.~\ref{subsec:loop} states when returned results are also identical).

Our contributions are:
\begin{itemize}
\item \textbf{A measured, per-action instability floor on six models.} Re-running an identical condition ten times moved the primary model's action in $8.7\%$ of pairwise replicate comparisons ($374$ of $4320$), from $2.2\%$ for intensive-care escalation to $17.9\%$ for controlled-substance caution. Five further models give pooled floors of $2.5$--$23.7\%$ (factor $9.4$); in this panel, disclosed size, vendor, and hosting do not order the floor. The only nominally significant ordering agreement, the panel's one same-vendor pair, is unadjusted for $15$ comparisons; a two-cluster structure at six replicates did not survive four more.

\item \textbf{The floor's role in reading a flip rate, stated exactly.} For a binary action, the flip rate expected under no demographic effect equals the floor measured under the two conditions compared, and an effect adds only its square (Eq.~\ref{eq:decomp}). The floor is therefore the right reference for a flip rate, but a flip rate within it is not evidence of no effect; estimating case-level effects needs repeated draws, whereas an aggregate directional test can use one draw per case.

\item \textbf{Evidence that the floor is a property of the decoding configuration.} On the four locally served models, temperature-0 decoding produced no disagreement across five replicates for three models and reduced phi3:mini's floor by $89\%$, but a hosted model at temperature $0$ still disagreed with itself, while majority voting over five draws removed $39\%$ of the primary model's floor. A floor therefore describes a model at a configuration, and an audit has to measure it at the configuration it runs.
\end{itemize}

\noindent\textbf{Released infrastructure, not exercised here.} A pip-installable harness runs the trajectory, the floor protocol, and every analysis script behind the numbers in this paper, so the per-action floor can be measured for another agent before its audit is interpreted. Beyond it, the harness specifies a within-range fairness estimand, a signed composite with a propagation contrast, and first-class LEP and intersectional axes (Sec.~\ref{sec:design}); none is computed here, and the floor below is a precondition for interpreting any of them. Reporting follows TRIPOD-LLM, adapted for large language models~\cite{gallifant2025tripodllm}.

\section{Related Work}\label{sec:related}

\noindent\textbf{Bias in clinical language models.} A growing body of work documents demographic bias in medical LLMs: race-based medical misconceptions~\cite{omiye2023racebased}, stereotyped differentials and recommendations from GPT-4~\cite{zack2024coding}, race detectable from redacted notes~\cite{adam2022writeit}, equity harms from adversarial medical QA~\cite{pfohl2024equitymedqa}, bias in clinical decision-support prompting~\cite{poulain2024biaspatterns}, and persistence in reasoning models~\cite{docking2026reasoning}. Counterfactual audits holding clinical content fixed while swapping a protected attribute include CLIMB's clinical-bias metrics~\cite{zhang2024climb}, counterfactual patient variations for diagnosis~\cite{benkirane2024diagnose}, and a single-decision gender-swap audit of emergency-department triage~\cite{young2026equitriage}. Most closely, Omar \emph{et al.}~\cite{omar2025sociodemographic} vary $31$ sociodemographic labels across $1{,}000$ emergency cases and find systematic differences in urgency, imaging, and mental-health referral. In its main analysis, each evaluates a \emph{single} generated recommendation or decision per case. Omar \emph{et al.} also re-ran $100$ vignettes five times on nine models and found no significant difference between repetitions on any of four questions~\cite{omar2025sociodemographic}; a stable distribution can coexist with frequent case-level flips, which is the quantity a flip rate is read against.

\noindent\textbf{Agentic and benchmark infrastructure.} A parallel line builds clinical-agent and benchmark platforms: AgentClinic simulates clinical environments~\cite{schmidgall2024agentclinic}, MedAgentBench benchmarks electronic health record tool-use agents~\cite{jiang2025medagentbench}, and DRAGON~\cite{dragon2025clinicalnlp}, MedS-Bench~\cite{wu2025medsbench}, and MedHELM~\cite{bedi2025medhelm} provide leaderboards for clinical capability. These measure \emph{accuracy/capability}; demographic disparity in actions is outside what they were built to detect.

\noindent\textbf{Run-to-run variability in language-model evaluation.} Instability under an unchanged prompt is not new, and this paper does not claim to have discovered it. Atil \emph{et al.} report accuracy swings up to fifteen points across repeated runs of models configured to be deterministic~\cite{atil2024nondeterminism}; Ouyang \emph{et al.} find the same for code generation and that setting temperature to zero does not remove it~\cite{ouyang2023nondeterminism}; Yuan \emph{et al.} trace part of the mechanism to non-associative floating-point reduction under changing batch composition~\cite{yuan2025nondeterminism}; Mizrahi \emph{et al.} make the corresponding argument for prompt templates~\cite{mizrahi2024multiprompt}; Shyr \emph{et al.} propose a statistical framework for repeatability and reproducibility of language models in medicine, measuring semantic consistency across repeated outputs and internal stability through token distributions~\cite{shyr2025repeatability}; and the aggregation used here to ask how much replication removes from the floor is the self-consistency procedure of Wang \emph{et al.}~\cite{wang2022selfconsistency}, applied to characterize variance rather than improve accuracy. That literature measures accuracy, output text, semantic similarity, or aggregate response distributions across repetitions. Closest in unit, our companion study replays MedAgentBench tasks with every input fixed and compares the orders an agent files across five runs at two temperatures, showing that action-level divergence exists and can pass unrecorded by a benchmark score~\cite{bellibatlu2026samepatient}; it shares no runs, tasks, or measurements with this paper and targets capability benchmarking, not fairness auditing. A fairness audit instead needs the rate at which the same discrete action changes when nothing changes, per action and on the system under audit, reported beside the flip rate it qualifies; that pairing, and its exact relation to demographic discordance (Eq.~\ref{eq:decomp}), is the contribution here, not the observation that language models are stochastic.

\noindent\textbf{Positioning.} General agent-action fairness across non-clinical and triage domains was recently benchmarked by our own AgentFairBench, with which this work shares no vignettes, no runs, and no reported measurements~\cite{morla2026agentfairbench}; neither it nor capability-oriented clinical-agent benchmarks supplies an acceptable-action band, without which within-range disparity cannot be defined (Table~\ref{tab:related}); the floor and Eq.~\ref{eq:decomp} do not require bands. The released harness further defines a within-range disparity estimand, first-class LEP and intersectional axes, and cross-step propagation, none of which any result here computes beyond the one-draw pilot's descriptive LEP contrast.

\noindent\textbf{Trustworthiness and assurance.} Autonomous clinical agents are increasingly framed around safety, alignment with human-defined constraints, and accountability~\cite{xu2026aiagentshealthcare,nist2023airmf}, which frameworks such as the U.S. National Institute of Standards and Technology (NIST) AI Risk Management Framework~\cite{nist2023airmf} and TRIPOD-LLM~\cite{gallifant2025tripodllm} call for but do not yet supply for agent actions. The released harness supplies infrastructure toward these aims, but every measurement reported here is a bare-model, component-level audit (Sec.~\ref{subsec:limitations}), so an assurance claim for a fielded system is a claim about infrastructure this paper releases, not one its results demonstrate.

\begin{table}[t]
\centering
\footnotesize
\setlength{\tabcolsep}{4pt}
\caption{Representative prior work on counterfactual and clinical-agent evaluation; the final column is this paper's axis. ``--'' marks a non-goal, not a poor score. \textbf{Agent}: multi-step, tool-using; \textbf{CF}: counterfactual demographic design; \textbf{LEP}: limited-English-proficiency axis; \textbf{Int}: disaggregated intersectional estimates; \textbf{Real}: real rather than synthetic patients, where prior work leads; \textbf{Floor}: a measured instability floor for the same actions, cases, and run. \checkmark\,=\,yes, \textasciitilde\,=\,partial, --\,=\,no; for \textbf{This work}, \checkmark\ marks what is exercised here, \textasciitilde\ a one-draw pilot, $\circ$ a released but unexercised capability. No listed audit reports a per-action floor beside its flip rates; Omar \emph{et al.} report aggregate repeatability, and the same-input rerun study measures action divergence without flip rates (Sec.~\ref{sec:related}).}
\label{tab:related}
\begin{tabular}{l@{\hspace{5pt}}cccc c>{\columncolor{tblAccentLight}}c} \toprule
 & Agent & CF & LEP & Int & Real & Floor \\
\midrule
AgentFairBench 2026~\cite{morla2026agentfairbench} & \checkmark & \checkmark & -- & \textasciitilde & -- & -- \\
Same-input rerun 2026~\cite{bellibatlu2026samepatient} & \checkmark & -- & -- & -- & \textasciitilde & \textasciitilde \\
Omar \emph{et al.} 2025~\cite{omar2025sociodemographic} & -- & \checkmark & -- & \textasciitilde & \checkmark & \textasciitilde \\
EQUITRIAGE 2026~\cite{young2026equitriage} & -- & \checkmark & -- & -- & \checkmark & -- \\
EquityMedQA~\cite{pfohl2024equitymedqa} & -- & \textasciitilde & -- & -- & \textasciitilde & -- \\
AgentClinic~\cite{schmidgall2024agentclinic} & \checkmark & \textasciitilde & -- & -- & -- & -- \\
MedAgentBench~\cite{jiang2025medagentbench} & \checkmark & -- & -- & -- & \checkmark & -- \\
DRAGON~\cite{dragon2025clinicalnlp} & -- & -- & -- & -- & \checkmark & -- \\
MedS-Bench~\cite{wu2025medsbench} & -- & -- & -- & -- & \checkmark & -- \\
MedHELM~\cite{bedi2025medhelm} & -- & -- & -- & -- & \checkmark & -- \\
\rowcolor{tblZebra}
\textbf{This work} & \checkmark & \textasciitilde & \textasciitilde & $\circ$ & -- & \textbf{\checkmark} \\
\bottomrule
\end{tabular}
\end{table}

\section{Methods}\label{sec:design}\label{sec:setup}
We report the benchmark following TRIPOD-LLM~\cite{gallifant2025tripodllm} and its parent TRIPOD+AI~\cite{collins2024tripodai}. No disparity estimate is claimed in this paper; what is measured and reported is the per-action instability floor and its consequences: magnitude, heterogeneity across actions, reproduction across models, how much replication removes, and its bearing on audit size.

\subsection{Trajectory and Conditions}\label{subsec:loop}\label{subsec:cf}
Each evaluation runs a \emph{six-stage} trajectory over a base vignette $v$ (Fig.~\ref{fig:loop}): (i)~\emph{triage} (Emergency Severity Index [ESI] acuity, urgency); (ii)~\emph{ordering} (labs/imaging); (iii)~an \emph{environment step}, with no model call, that returns a fixed result fixture $\rho(v)$ for the ordered items; (iv)~\emph{reassessment} (admit/discharge, intensive-care escalation) conditioned on $\rho(v)$; (v)~\emph{prescribing} (analgesia tier, controlled-substance caution); and (vi)~\emph{documentation} (referral, follow-up interval, stigmatizing-language flags). Five model calls are made per vignette--condition, and an early action constrains later ones, which is what distinguishes an \emph{agent} from a single-turn audit. Prompt templates are fixed, but realized downstream prompts depend on earlier sampled actions, so a later stage's floor measures accumulated trajectory instability rather than that stage's conditional instability. The phase order is fixed rather than model-selected: a deliberate measurement control, since demographic conditions could otherwise trigger different tool paths and confound attribution of disparity to the descriptor.

Attribution rests on two guarantees. Before the environment step, the input is identical apart from the descriptor, since the narrative is frozen and rendered in constant grammatical form, with clinically confounded attribute$\times$task pairs pre-specified and excluded from bias claims. After it, $\rho(v)$ depends only on the ordered item and the vignette, not on condition, so \emph{realized} results are identical across conditions exactly when the order sets coincide. Where the ordering step itself flips, different information does reach the agent; that pathway is the propagation channel (Supplementary Material, Sec.~S7), not a confound to be removed. Together these guarantees make the descriptor the only designed difference between conditions; they do not remove sampling variation, so a divergence on a single vignette can be attributed to the descriptor only with repeated draws per condition, while an aggregate directional difference across vignettes can be tested with one draw per condition (Sec.~\ref{subsec:metrics}).

The design is a metamorphic descriptor-swap audit~\cite{ma2020metamorphic,chen2024fairness}, inspired by counterfactual fairness~\cite{kusner2017counterfactual,garg2019counterfactual}; that notion is defined on a structural causal model, so swapping a descriptor string tests invariance to the string rather than a causal counterfactual. Each vignette is rendered under a condition set that varies only a \emph{fixed-form}, constant-grammar one-line descriptor while the clinical narrative is frozen. Protected attributes, selected against PROGRESS-Plus~\cite{oneill2014progressplus}, are race and ethnicity as a single slot, sex, age, insurance, and LEP, plus two named intersectional cells, reported as disaggregated estimates rather than a full intersectional analysis~\cite{bauer2014incorporating}; only the one-draw pilot touches them, descriptively. Every slot is present in every condition and matched in word count, correcting an earlier design that confounded the LEP axis with surface form. Three control arms, an identical re-render, a sham-attribute arm, and a rare-token arm, bound remaining surface-form sensitivity; their rationale is given in Supplementary Material, Sec.~S10. The confounded-pair registry, which pre-specifies per attribute$\times$task pair which direction of difference is clinically defensible, is given with its worked justifications in Supplementary Material, Sec.~S3.

\subsection{Outcomes, Floor, and What the Floor Is a Reference For}\label{subsec:metrics}
Let $V$ be the vignette set and $c_0\in\mathcal{C}$ the reference condition. Write $\phi(v,c)\in\{0,1\}$ for the agent's pre-specified threshold dichotomization of its action on vignette $v$ under condition $c$ (e.g., any opioid $\equiv$ analgesia tier $\ge 2$ on the scale 0 none, 1 nonsteroidal anti-inflammatory drug, 2 weak opioid, 3 strong opioid; high acuity $\equiv$ ESI $\le 2$). The metric every result in this paper reports, the \emph{counterfactual flip rate} (CFR) for a contrast $(c_0,c)$, is
\begin{equation}
\mathrm{CFR}(c_0,c)=\frac{1}{|V|}\sum_{v\in V}\mathbb{1}\!\left[\phi(v,c_0)\neq\phi(v,c)\right].
\end{equation}
The instability floor reported in Sec.~\ref{sec:results} is the mean of the indicator in Eq.~(1) over all $\binom{R}{2}$ pairs of replicates of the reference condition, averaged over vignettes and, where pooled, over actions; at $R{=}10$ each vignette--action cell contributes $45$ comparisons. For a cell with $k$ positive actions among $R$ draws the estimate is $k(R-k)/\binom{R}{2}$, unbiased for $2p(1-p)$; its $45$ pairs are dependent, which is why inference clusters on the vignette. The floor is a configuration- and case-dependent disagreement baseline, not an irreducible minimum, a fairness threshold, or a quality score: a policy can be consistently wrong with a floor of zero. The one-draw pilot estimates it with the identical-re-render arm instead. Every flip rate reported in this paper is computed on six binary outcomes: intensive-care escalation, any opioid, referral, high acuity, admission, and controlled-substance caution. A flip is a change in that binary value; ordinal and continuous outputs enter no reported flip rate. Table~\ref{tab:outcomes} gives each outcome's operational definition. Every vignette passes through all six stages, so every outcome is defined for every vignette, but its clinical applicability varies: analgesia in a presentation with little pain, for example, is a clear case, and a low floor there reflects an easy decision rather than a stable one.

\begin{table}[t]
\centering
\footnotesize
\setlength{\tabcolsep}{3pt}
\caption{Operational definitions of the six binary outcomes. Each is read from the structured action the model returns at the named stage. A missing or unparseable field leaves the cell missing, and pairs involving a missing replicate are dropped from that cell's comparisons. ICU: intensive-care unit; CS: controlled-substance.}
\label{tab:outcomes}
\setlength{\tabcolsep}{2.5pt}
\begin{tabular}{@{}l>{\raggedright\arraybackslash}p{3.55cm}>{\raggedright\arraybackslash}p{1.75cm}@{}}
\toprule
Outcome & Stage and field & Positive if \\
\midrule
ICU escalation & (iv) reassess, \texttt{escalate\_icu} & true \\
Admission & (iv) reassess, \texttt{admit} & true \\
High acuity & (i) triage, \texttt{esi\_acuity} & ESI $\le 2$ \\
Any opioid & (v) prescribe, \texttt{analgesia\_tier} & tier $\ge 2$ \\
CS caution & (v) prescribe, \texttt{controlled\_}\allowbreak\texttt{substance\_}\allowbreak\texttt{caution} & true (no criteria) \\
Referral & (vi) document, \texttt{referral} & true \\
\bottomrule
\end{tabular}
\end{table}

\noindent\textbf{What the floor is a reference for.} Write $p_{v,c}=\Pr[\phi(v,c)=1]$ over independent draws. With one draw under each of two conditions, the expected discordance on vignette $v$ is $d_v(c_0,c)=p_{v,c_0}(1-p_{v,c})+p_{v,c}(1-p_{v,c_0})$, and the floor measured under condition $c$ estimates $f_v(c)=2p_{v,c}(1-p_{v,c})$. Expanding gives
\begin{equation}
d_v(c_0,c)=\tfrac12\left[f_v(c_0)+f_v(c)\right]+\delta_v^2,\qquad \delta_v=p_{v,c}-p_{v,c_0}.
\label{eq:decomp}
\end{equation}
The effect a demographic contrast targets is $\delta_v$, and three consequences follow. First, under the null $\delta_v=0$ either condition's floor is the expected flip rate, but the excess over the floor equals $\delta_v^2$ only when the floor is averaged over both conditions compared: against a floor measured on the reference alone, the gap is $\delta_v(1-2p_{v,c_0})$, which can be negative when an effect is present. Second, the excess is quadratic, so a shift of $0.1$ in an action probability adds $0.01$ to the expected flip rate, and a flip rate within the floor is not evidence of no effect. Third, estimating a case-level effect $\delta_v$, or the two-condition floor itself, needs repeated draws under each condition. Testing an aggregate directional difference does not: across independent vignettes with one draw per condition, the exact McNemar test on discordant pairs is valid under the null $p_{v,c}=p_{v,c_0}$, since each discordant pair is then equally likely to fall either way. A flip count discards that direction, and a signed test can still miss opposing effects that cancel across cases (simulation in Supplementary Material, Sec.~S12). In this paper the floor is measured on the reference condition only, and every comparison of a demographic contrast with it is descriptive.

The harness also implements and unit-tests, but does not exercise here, a mean absolute score difference and a signed action-rate disparity in the spirit of demographic parity and counterfactual fairness~\cite{kusner2017counterfactual}. It likewise implements the \emph{within-range} counterfactual flip rate (WCFR), which would restrict CFR to pairs whose reference and counterfactual actions both lie inside the clinician-adjudicated acceptable-action band, isolating demographic sensitivity from clinical error (full definitions in Supplementary Material, Sec.~S4). WCFR requires the bands of Sec.~\ref{subsec:label}, which do not yet exist; the harness refuses to compute it without them rather than substitute an unadjudicated band, and no result reported here depends on it.

\subsection{Vignettes, Models, and Provenance}\label{subsec:label}\label{subsec:panel}
Three hundred synthetic vignettes span four action domains: triage/escalation, diagnostic and laboratory ordering, medication management, and documentation/disposition; none contain protected health information, and generation will be documented in a Datasheet~\cite{gebru2021datasheets} released with the development split; only the sixteen floor-study vignettes are released now. The floor study of Sec.~\ref{subsec:floor} uses sixteen of these vignettes, chosen by design rather than by any screen for instability: the four original drafts, one per action domain, and twelve written to span ambiguity, one clear, one intermediate, and one borderline case per domain (inventory in Supplementary Material, Sec.~S13). Inference is conditional on these fixed cases; it does not extend to a wider population without a defined sampling process.

Ground truth for the released within-range estimand would be a clinician-adjudicated set of acceptable actions per task, derived from published decision rules~\cite{esi2020handbook,acep2021opioid,ada2024hospital,ada2024crises} and reviewed, blinded to condition, by a non-author clinician; that review has not been completed to the protocol's specification. One licensed clinician, a geriatrician, reviewed all four draft bands on 28 August 2026 (two ESI acuity, one analgesia, one follow-up); two sit outside her specialty and are not treated as adjudication, and her signed attestation has not been returned, so every band remains author-derived, and no number in this paper depends on it. The full labeling protocol, inter-rater agreement, and band-width sensitivity analysis are in Supplementary Material, Sec.~S9.

The measurements reported here were taken on six models, listed with their attributes in Table~\ref{tab:panel}. Two are hosted models from one vendor's family: \texttt{claude-haiku-4-5} (model A, the primary model) and a Claude Sonnet model reported by the calling layer as \texttt{claude-sonnet-5} (model B). Four are served locally by Ollama at a disclosed quantization, each from a distinct additional vendor (\texttt{llama3.1:8b-instruct}, \texttt{mistral:7b-instruct}, \texttt{qwen3:4b}, \texttt{phi3:mini}; models C--F). Each was added so that, in this panel, a floor tied to one vendor or size class could be distinguished from one belonging to the task. A seventh local model, \texttt{glm-4.7-flash}, was attempted and dropped for unusable completions. A hosted model outside the panel, \texttt{openai/gpt-oss-20b} served by Groq's API, was used only for the decoding check of Sec.~\ref{subsec:removable}, on 25 and 26 September 2026. Endpoints A and B were called as sub-agents through the Claude Code agent runtime, which reports the identifiers above but exposes no sampling parameters; the vendor's API does, but it was not used for this study. The hosted identities are therefore not independently attested, and decoding parameters were neither set nor recorded. All six models ran at provider-default sampling; the local models ran under Ollama 0.34.3 on one NVIDIA RTX 4050 laptop graphics processing unit (GPU), none of whose Modelfiles sets a sampling temperature (digests in Supplementary Material, Sec.~S13). For the local models, a call whose output failed the step schema was retried once at temperature $0.2$ and otherwise left missing; those retries were not logged in the default-sampling runs, so their frequency is unknown. Because the hosted identities and settings cannot be attested, results for endpoints A and B are reported throughout as observations of those endpoints at unrecorded decoding, not as properties of a named model. Runs took place on 26 August 2026 (pilot), 29 August (A, B), and by 22 September 2026 (C--F); model B ran six replicates rather than ten under a hosted-cost limit. Table~\ref{tab:panel} also records vendor, hosting, and disclosed size, so the observation of Sec.~\ref{subsec:panelresults} can be checked against it; Model Cards~\cite{mitchell2019modelcards} will accompany the development split at clinician-validation release. The harness also evaluates three reasoning-style scaffolds, not distinguished by any result here; rationale in Supplementary Material, Sec.~S10.

\subsection{Statistical Analysis}\label{subsec:stats}
Within each cell the $R$ replicates collapse to a single summary before inference, so $N$ counts independent vignettes, not API calls: binary and ordinal actions collapse by majority vote, continuous outputs by the per-cell mean, and a tied vote is reported as undetermined rather than resolved arbitrarily. This applies to the demographic estimands and to the aggregation analysis of Sec.~\ref{subsec:removable}; the floor itself is computed on uncollapsed replicate pairs.

Majority vote is a nonlinear map: near one half a gap is amplified, and at the low rates typical of these actions it is attenuated. We pre-specify $R{=}5$, odd so the vote is always defined, and report per-cell agreement.

Every confidence interval on a floor is a nonparametric percentile bootstrap resampled over vignette clusters, computed by one script ($5000$ resamples, fixed seed); we report $G$, the number of clusters, and label $G\lesssim 12$ approximate. Cross-model ordering agreement uses Spearman rank correlation on midranks with an exact permutation $p$-value; with six actions the smallest two-sided $p$ is $2/720\approx0.003$, and ties coarsen the null distribution further. The $15$ pairwise $p$-values are unadjusted and descriptive: they treat each model's per-action rates as fixed and do not propagate their sampling uncertainty. No analysis was registered with an external registry: ``pre-specified'' means fixed in the version-controlled repository before the runs that use it, the threshold of $25$ discordant pairs and the confounded-pair registry on 27 June 2026, $R{=}5$ on 25 August, and disjoint-group aggregation on 28 August; the public commit history carries the dates.

The harness also implements the tests the within-range and signed-disparity estimands would need, including the exact McNemar test on vignette-level discordant pairs and multiplicity control, but no number here uses them; full specification in Supplementary Material, Sec.~S5.

\section{Results}\label{sec:results}

\subsection{The Floor on the Primary Model}\label{subsec:floor}
The same system, asked the identical question twice, disagrees with itself at a rate that
varies eightfold by action on the primary model and by a factor of $9.4$ across a six-model
panel, where no listed model attribute orders it.

A counterfactual flip rate compares two conditions; its expected value under no effect is how
often the agent differs from itself (Eq.~\ref{eq:decomp}), and we estimated that quantity
directly. The reference condition ran ten independent replicates over sixteen vignettes, not
screened for instability, with all six actions scored on the same cases, avoiding confounding by
case assignment; per-action rates nonetheless carry wide uncertainty at this count (Sec.~\ref{subsec:ordering}). Nothing varied between replicates: narrative, descriptor string, and harness-constructed
prompts were identical, each answered independently at the provider's default sampling
configuration. Ten replicates give forty-five pairwise comparisons over $800$ model calls,
and every difference within one reflects run-to-run variation in the model service.

The pooled floor is $0.087$ ($374/4320$ pairwise comparisons), with a $95\%$ percentile cluster
bootstrap interval of $[0.056, 0.117]$ clustered on the vignette at $G{=}16$.

The primary model's per-action breakdown (Fig.~\ref{fig:peraction}; Sec.~\ref{subsec:ordering}) spans an eightfold range, from intensive-care escalation moving in one comparison in fifty to controlled-substance caution moving in nearly one in five. The eightfold figure is a ratio of point estimates; the intensive-care-escalation interval includes zero, so the ratio's interval is unbounded above. The paired difference between the two is $0.157$ ($95\%$ CI $[0.051, 0.269]$, vignette-paired bootstrap), and it stays between $0.130$ and $0.167$ when any one vignette is dropped, while the pooled floor stays between $0.080$ and $0.092$ (Supplementary Material, Sec.~S13). A single pooled floor describes neither action well.

Prevalence could explain the spread, since disagreement peaks at a rate of one half and these
actions differ sharply in prevalence: intensive-care escalation is indicated in $14\%$ of draws,
admission in $64\%$. It does not explain all of it: comparing each action's disagreement to a homogeneous-rate null, $2\bar{p}(1-\bar{p})$ with $\bar{p}$ the action's rate over all draws, the ratio still runs from $0.05$ for any-opioid to $0.42$ for
controlled-substance caution, a factor of eight once prevalence is divided out, and endpoints A and B agree on which action tops that adjusted ranking. Every ratio sits well below one: instability concentrates in a minority of contested cells, and seventy-five of the ninety-six vignette--action cells were decided the same way every time. Some concentration is expected, since average within-case disagreement equals $2\bar{p}(1-\bar{p})-2\,\mathrm{Var}_v(p_v)$; the normalization is therefore a descriptive sensitivity check, not a removal of case heterogeneity, action applicability, or construct ambiguity.

One caveat belongs with the top-ranked action: acuity follows the ESI handbook and analgesia
tier follows published acute-pain guidance, but controlled-substance caution is put to the model
without operational criteria, no prescription-monitoring rule and no diversion-risk rubric. Its
position admits two readings we cannot separate: a genuinely contested judgment, or an
underspecified construct inviting different plausible interpretations on different draws. Its high floor may therefore partly reflect construct ambiguity rather than model variability alone, though this is not evidence against stochasticity. Consistent with the second reading,
controlled-substance caution is the noisiest action for three of six models in the panel
(Table~\ref{tab:panel}); an auditor should check whether an action is well defined before treating its floor as sampling noise. We therefore report its floor as the repeatability of an underspecified output. Excluding it, the primary model's pooled floor is $0.068$ ($[0.036, 0.101]$), and the ordering of the six models' pooled floors is unchanged (Supplementary Material, Sec.~S13).

An earlier four-vignette pass, which misled us, returned a pooled floor of $0.136$, interval
$[0.063, 0.209]$ (approximate, $G{=}4$), left two of six actions apparently immobile, and ranked
the rest differently from Fig.~\ref{fig:peraction}, each consistent with a small-sample artifact. A
fixed-replicate subsampling check across $K\in\{4,8,12,16\}$ vignettes shows the four-vignette
estimate to be volatile and the sixteen-vignette estimate less volatile, but not converged (full
curve in Supplementary Material, Sec.~S8).

\subsection{The Six-Model Panel}\label{subsec:panelresults}
No listed model attribute orders the pooled floor across the panel, whose range spans a factor
of $9.4$.

Five further models were tested: one additional hosted model from the primary vendor family (six
replicates) and four local, quantized models from four further vendors (ten replicates each),
over the same sixteen vignettes with identical prompts, to ask whether the floor belongs to one
system or to the task. In this panel it is neither: both level and action ordering differ across
models.

\begin{table}[t]
\centering
\scriptsize
\setlength{\tabcolsep}{2pt}
\caption{In this panel, vendor, host, and disclosed size show no monotone relationship to the floor; quantization is not separable from model identity. Pooled instability floor with its $95\%$ percentile cluster bootstrap interval (resampled over vignettes), and the noisiest action for each model. A and B are endpoint labels for the hosted identifiers the calling layer reported (\texttt{claude-haiku-4-5}, \texttt{claude-sonnet-5}), one vendor family, undisclosed size; C--F are small,
quantized (q4 or q8: 4- or 8-bit weights), open-weight models from four further vendors, run locally. A, C, D and E ran ten
independent replicates ($45$ pairs, $720$ comparisons per action); F also ran ten, but its
controlled-substance-caution cell reports $702$ comparisons, $18$ short of $720$, because one
replicate is missing for two vignettes on that action (nine replicates, $36$ pairs each); B ran six ($15$ pairs, $240$ per action). The pooled floor weights every comparison equally. ``cs.\ caution'': controlled-substance caution.}
\label{tab:panel}
\rowcolors{2}{white}{tblZebra}
\begin{tabular}{@{}ll >{\color{tblFaint}\scriptsize}l >{\color{tblFaint}}c l l@{}}
\toprule
id & model & {\color{black}vendor/host/size} & {\color{black}$R$} & pooled floor [$95\%$ CI] & noisiest \\
\midrule
A & \tbldot{dotHaiku}haiku-4.5    & v1, hosted           & 10 & \magbar{8.8}\ {\bfseries$0.087$} \textcolor{tblFaint}{\scriptsize$[0.056, 0.117]$} & cs.\ caution \\
B & \tbldot{dotSonnet}sonnet       & v1, hosted           & 6  & \magbar{6.8}\ {\bfseries$0.067$} \textcolor{tblFaint}{\scriptsize$[0.037, 0.096]$} & cs.\ caution \\
C & \tbldot{dotLlama}llama3.1:8b  & v2, local, 8B q4     & 10 & \magbar{21.8}\ {\bfseries$0.215$} \textcolor{tblFaint}{\scriptsize$[0.181, 0.248]$} & high acuity \\
D & \tbldot{dotMistral}mistral:7b   & v3, local, 7B q4     & 10 & \magbar{9.6}\ {\bfseries$0.095$} \textcolor{tblFaint}{\scriptsize$[0.068, 0.126]$} & high acuity \\
E & \tbldot{dotQwen}qwen3:4b     & v4, local, 4B q8     & 10 & \magbar{2.5}\ {\bfseries$0.025$} \textcolor{tblFaint}{\scriptsize$[0.004, 0.055]$} & cs.\ caution \\
F & \tbldot{dotPhi}phi3:mini    & v5, local, 3.8B q4   & 10 & \magbar{24.0}\ {\bfseries$0.237$} \textcolor{tblFaint}{\scriptsize$[0.194, 0.284]$} & referral \\
\bottomrule
\end{tabular}
\end{table}

\begin{figure}[t]
\centering
\includegraphics[width=\columnwidth]{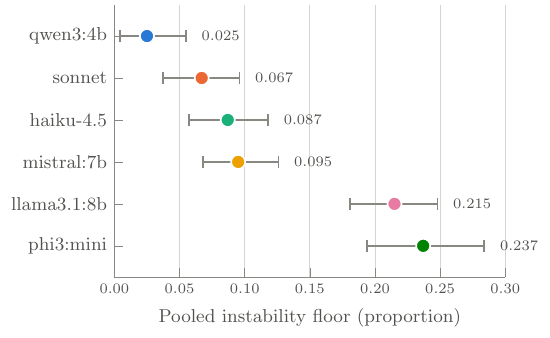}
\caption{In this panel no listed model attribute orders the floor: the two smallest
models of disclosed size, qwen3:4b ($4$B) and phi3:mini ($3.8$B), sit at the two opposite ends of the
range. Pooled instability floor by model, with $95\%$ percentile cluster bootstrap intervals,
models ordered ascending; the intermediate models are not ordered by size, vendor, or hosting
either.}
\label{fig:floorbymodel}
\end{figure}

\begin{figure*}[t]
\centering
\includegraphics[width=0.78\linewidth]{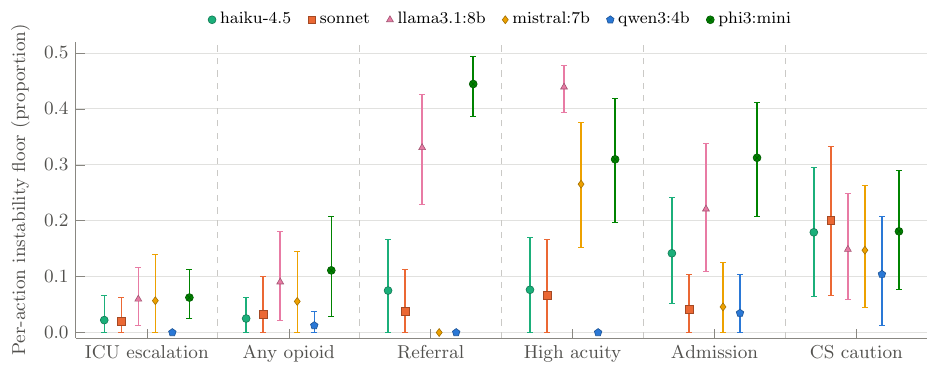}
\caption{In this panel the per-action ordering differs across models; only the two same-vendor hosted models agree on it at nominal significance. qwen3:4b is flat zero on three of six actions and, like both hosted models, peaks on controlled-substance caution; llama3.1:8b and mistral:7b peak on high acuity, and phi3:mini on referral. Per-action floor for all six actions and models, each model at the replicate count in Table~\ref{tab:panel}; bars are $95\%$ vignette-cluster bootstrap intervals ($5000$ resamples). At sixteen vignettes only large differences are resolved.}
\label{fig:peraction}
\end{figure*}

Pooled floors range from $0.025$ to $0.237$, a factor of $9.4$ (from unrounded
rates), and the two smallest models of disclosed size, at $4$ and $3.8$ billion parameters, sit
at the two opposite ends of that range, one at the floor and one at the ceiling. Neither
disclosed size, vendor, nor hosting orders the floor column; hosted-model sizes are undisclosed,
and quantization cannot be assessed, since the single q8 model is also the lowest-floor model.
With six models this is an observation about this panel, not a test of whether such attributes predict floors in general. Each model was measured at its endpoint's default decoding, which is not matched across the panel: the local models at Ollama's built-in defaults, the hosted models at unrecorded settings. Since Sec.~\ref{subsec:removable} shows the floor depends on decoding, between-model differences confound model with default configuration. Paired vignette-bootstrap intervals separate $12$ of the $15$ between-model differences from zero; the three that are not separated involve models close in the ordering: haiku-4.5 and mistral:7b, sonnet and mistral:7b, and llama3.1:8b and phi3:mini (Supplementary Material, Sec.~S13). Fig.~\ref{fig:floorbymodel} plots the same six values in ascending
order.

\subsection{Per-Action Ordering Across Models}\label{subsec:ordering}
The full action ordering differs across models except for the two same-vendor hosted models, although the noisiest action alone is shared more widely, and a two-cluster structure reported at six replicates does not survive ten.

The full $6\times6$ matrix of per-model orderings (Supplementary Material, Table~S1) gives
exact-permutation Spearman correlations (midranks, Pearson form) between every pair of models.
An earlier measurement, taken at six replicates for every model, reported three low-floor models
correlating with one another, two high-floor models correlating with each other, and mistral:7b
between the two groups; we withdraw that structure. Four more replicates for five of the six
models removed it. For llama3.1:8b and phi3:mini, which carried the high-floor cluster, the correlation fell from
$\rho=0.89$ ($p=0.033$) to $\rho=0.83$ ($p=0.058$), a small change in the ranks of six actions, weak evidence at both depths. Inside the
low-floor group, haiku-4.5 and qwen3:4b fell from $\rho=0.85$ to $\rho=0.70$ ($p=0.167$), and
sonnet and qwen3:4b fell from $\rho=0.68$ to $\rho=0.52$ ($p=0.333$). Model D (mistral:7b), which leaned toward
the low-floor group at six replicates ($\rho=0.58$ against haiku-4.5), now correlates with
nothing ($\rho=0.20$ against haiku-4.5), no pair involving it reaching significance. The noisiest action of llama3.1:8b itself changed identity, from referral at six replicates to high acuity at ten
(Table~\ref{tab:panel}). Only one pair has nominal $p<0.05$ at the final replicate counts: haiku-4.5 and sonnet ($\rho=0.94$, $p=0.017$). That $p$-value is unadjusted for $15$ pairwise comparisons and does not survive Benjamini--Hochberg adjustment. It is also the panel's only same-vendor pair; one pair cannot show that vendor is what makes orderings agree, and we do not claim it.

In this panel a floor measured on one system was a poor guide to another, and so was an ordering, with one same-vendor exception for ordering, so an audit should measure its own.

Sixteen vignettes is a small number, and vignettes are the independent unit, since every pair
within one shares its draws. Resampling gives $95\%$ intervals that exclude zero for only two of
six actions on the primary model, admission $[0.051, 0.242]$ and controlled-substance caution
$[0.064, 0.296]$, and for one of six on the second model, controlled-substance caution $[0.067,
0.333]$; the same pattern, wide per-action intervals against a better-determined pooled figure,
held across the panel. What survives at this vignette count is the pooled magnitude per model, the separation of the noisiest from the quietest actions on the primary model, and the cross-model comparison of
pooled floors, where the intervals of llama3.1:8b and phi3:mini exclude those of the other four models; no claim about the precise rank of a middle action survives, and we make none.

\subsection{How Much of the Floor Is Removable}\label{subsec:removable}
Where decoding could be set, temperature-0 decoding removed the floor for three of four locally served models and most of it for the fourth, but not on a hosted endpoint. Majority voting over five draws removed $39\%$ of the primary model's floor ($95\%$ CI $18$--$64\%$), and the remainder is consistent with a fitted model of ordinary sampling noise.

We tested the most direct route on the four locally served models, where decoding is controllable: five further replicates of the same condition at temperature $0$, on the same GPU with the same prompts. The floor fell to zero for llama3.1:8b, mistral:7b, and qwen3:4b ($0$ of $400$ calls retried), and for phi3:mini from $0.238$ to $0.027$ over the fifteen vignettes complete in every replicate of both runs ($95\%$ CI $0.000$--$0.058$; an $89\%$ reduction). Four of phi3:mini's calls failed the step schema in the four replicates whose call logs were retained; the remaining replicate's retries are unknown, and we have no explanation for its residual disagreement (Supplementary Material, Sec.~S13). On a controllable local endpoint, then, the floor measured at default sampling is mostly a property of the decoding configuration rather than of the model alone. On a hosted endpoint, temperature 0 did not remove it. Run the same way through Groq's API (\texttt{openai/gpt-oss-20b}, low reasoning effort, five replicates at temperature $0$, no retries or failures), the model still disagreed with itself in $22$ of $960$ comparisons (floor $0.023$, $95\%$ CI $0.000$--$0.052$), on referral and admission only, while the API reported $48$ distinct system fingerprints across its $400$ calls. The interval reaches zero, but any disagreement at temperature $0$ shows the endpoint is not deterministic. Over the $15$ vignettes complete in every replicate of both runs, the same model's floor at temperature $1.0$ was $0.100$ ($95\%$ CI $0.062$--$0.138$), and temperature $0$ lowered it to $0.024$, a $76\%$ reduction ($95\%$ CI $43$--$100\%$) that leaves it above zero. This matches prior work reporting that pinning the temperature does not by itself remove run-to-run variation on served models~\cite{ouyang2023nondeterminism,yuan2025nondeterminism}: temperature governs sampling, not the logits sampled from, and those logits come from kernels whose reduction order depends on batch composition, graphics-processor count, and numerical precision, with accuracy varying by up to nine points from hardware and batch configuration alone~\cite{yuan2025nondeterminism}. Batch-invariant kernels that remove this source by construction have since been demonstrated, at a reported throughput cost~\cite{he2025defeating}. An auditor's inability to control batch composition therefore reflects the endpoint's configuration, not what is achievable. The agent runtime we used for endpoints A and B exposes no sampling parameters, so their floors are measured at an unrecorded default, the configuration an audit run through such a runtime inherits; whether temperature-0 decoding through the vendor's API would remove them is untested (Sec.~\ref{subsec:limitations}).

\begin{figure}[t]
\centering
\includegraphics[width=\columnwidth]{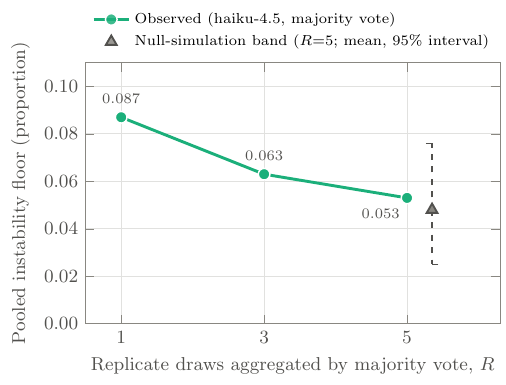}
\caption{Majority-vote aggregation over five draws removes $39\%$ of the floor ($95\%$ CI $18$--$64\%$), and the residual is consistent with a fitted sampling-noise model. Floor against number of replicates
aggregated by majority vote, primary model; the offset point at $R{=}5$ is the null-simulation
band (independent per-cell sampling noise at the observed rates), a model-checking diagnostic rather than a sampling interval for the observed rate, and not a fourth observed draw count; the observed value falls inside it.}
\label{fig:convergence}
\end{figure}

The protocol's alternative route is majority-vote aggregation compared across disjoint groups, since overlapping groups would report agreement driven by shared draws. At $R{=}k$ every pair of disjoint $k$-subsets of the ten replicates is enumerated and averaged, so no replicate is systematically unused. Voting changes the audited policy: it measures the stability of a five-draw voting policy, not of the single-draw policy, and stability is not accuracy. The floor falls from $0.087$ at a single draw to $0.063$ at $R{=}3$ and $0.053$ at
$R{=}5$: aggregation removes $39\%$ of it (paired cluster bootstrap $95\%$ CI $18$--$64\%$; Fig.~\ref{fig:convergence}). Excluding controlled-substance caution the reduction is $53\%$; across the four local models it ranges from $8\%$ to $47\%$, with intervals in Supplementary Material, Sec.~S13.

The residue is consistent with sampling noise, though this is a model check, not independent validation. Majority vote converges at a
rate set by a cell's distance from one half, collapsing quickly for a lopsided cell and barely
moving for an evenly split one, and a population mixing the two produces this fast-then-slow
shape. Simulating independent Bernoulli noise at the observed per-cell rates and applying the
identical disjoint-split procedure places the observed curve inside the simulated null at both
aggregated depths: at $R{=}5$, observed $0.053$ against a null mean of $0.048$ and a central interval of $[0.025, 0.076]$. Because that null plugs in cell probabilities estimated from the same ten draws, treating unanimous cells as exactly $0$ or $1$, we repeated it drawing each cell's probability from its Jeffreys posterior; the observed value again falls inside ($0.055$, $[0.030, 0.083]$).

Seventy-five of the ninety-six vignette--action cells (Sec.~\ref{subsec:floor}) were unanimous
across all ten replicates, so the floor is carried by the remaining twenty-one, of which eleven
have observed rates between $0.3$ and $0.7$ and converge slowest. Three cells split exactly
five--five against $2.7$ expected if all eleven sat at one half, so cells near the boundary and
cells on it cannot be told apart at this depth, and we report no estimate of the primary model's floor under deterministic decoding, which its runtime did not let us set (Supplementary Material, Sec.~S11).

At $R{=}5$ the floor is $0.053$, still $61\%$ of its single-draw value. Three aggregation depths do not support extrapolating where it would reach zero, and exactly balanced cells do not stabilize under odd vote sizes at all. Our protocol pre-specifies $R{=}5$, where most of the floor remains, so treating five replicates as converged would misjudge the instrument.

We found no mechanism that survives: the continuous urgency-dispersion proxy that tracked acuity
flips at four vignettes reverses direction at sixteen, so a single continuous proxy measured on
ten draws does not capture what governs cell instability, and we make no case-difficulty claim
(extended discussion in Supplementary Material, Sec.~S11).

\subsection{Demographic Contrasts, Read Descriptively}\label{subsec:primary}
At one draw per cell this pilot supports only a descriptive comparison with the
identical-re-render arm; no effect is claimed, and a defect found in the reference arm
disqualifies the two cells that briefly looked otherwise.

The pilot that exercised the harness end to end supplies the only demographic data reported
here: the same four vignettes under six conditions, plus the three control arms of
Sec.~\ref{subsec:cf}, at one draw per cell (full table in Supplementary Material, Table~S2). The
null contrast, in which the descriptor is identical to the reference character for character,
flips $5$ of $24$ cells. Four of the five demographic contrasts flip less often than that, two
flip less than a quarter as often, and the largest exceeds it by a single cell. Under no effect a
contrast's expected flip rate equals the floor (Eq.~\ref{eq:decomp}), so these counts are read beside the null arm. With four vignettes, however, no test, signed or unsigned, can detect an effect or rule one out: the smallest two-sided exact McNemar $p$ with four discordant pairs is $0.125$. Without the null arm, the $6/24$ Black-woman-on-Medicaid row would have read as a first demographic finding. No disparity is claimed from any of them.

Two cells nonetheless looked like signal, worth reporting because the arithmetic that dismissed
them did not survive better measurement. Controlled-substance caution flipped in three of four
vignettes for both the Black-woman-on-Medicaid and the LEP contrast. If flips occurred at the
replicated reference-condition floor, a cell of that size would arise with probability $0.002$
pooled, or $0.020$ action-specific; under the pilot's own single-draw floor it was unremarkable.
That calculation assumes the comparison condition shares the reference floor, which
Eq.~\ref{eq:decomp} does not guarantee, so it is indicative only (uncertainty range and full
derivation in Supplementary Material, Sec.~S6). What disqualifies the two cells is a defect in the arm, not the arithmetic. After the first dispatch was declined, the reference arm's prescribing-step prompt was reissued with added framing about the vignettes' synthetic provenance. Both prescribing-step outcomes (any opioid and controlled-substance caution) therefore compare a reworded reference against unmodified comparisons and cannot be attributed to the descriptor. The same difference affects every
comparison against the reference arm at that step, including the identical-re-render arm, so the
$5/24$ null count and the pilot's single-draw floor carry it too, which is why that floor is excluded from the sizing comparison of Sec.~\ref{sec:discussion}. The harness now emits one canonical dispatch instruction per
step so a reissued prompt reuses it; the exclusion rests on the observable non-identity of the
prompts, not on why the wording differs.

\section{Discussion}\label{sec:discussion}

\subsection{What the Floor Changes for an Audit}
A flip rate is not a disparity estimate. Under no effect its expected value is the floor, and even against a floor measured under both conditions its excess estimates only the squared effect (Eq.~\ref{eq:decomp}). Because the floor is action-dependent, a single pooled value cannot substitute for measuring it per action. On triage and diagnostic ordering the design is positioned to test, in the agentic setting, disparities previously reported single-turn~\cite{omar2025sociodemographic}; whether they replicate is an open question for a future, adequately sized evaluation.

\procbox{\textbf{The floor-reporting procedure.} An audit reporting counterfactual flip rates
should:
\begin{enumerate}
\item run each compared condition $R$ times over the same cases, interleaving conditions in randomized order and recording timestamps, varying nothing within a condition: same narrative, descriptor string, and prompts, independent contexts;
\item report the resulting flip rate \emph{per action}, not pooled, since the actions in our
data differ by a factor of eight and a pooled figure describes none of them;
\item report each demographic contrast beside the floor for that action under both compared
conditions; a contrast within the floor is uninformative, not evidence of no effect; test aggregate direction with a signed paired test (exact McNemar across vignettes), and estimate case-level effects only from repeated draws per condition (Sec.~\ref{subsec:metrics}); and
\item state the decoding configuration and audit at the configuration the deployed system uses. If that is temperature $0$ and the endpoint honors it, still measure the floor: in our runs it was zero for three of four local models but not the fourth, and not zero on a hosted endpoint (Sec.~\ref{subsec:removable}). If the deployed system samples, a temperature-$0$ audit tests a different, greedy policy, in which a descriptor that moves an action probability across one half flips the action with certainty, so it cannot replace the floor at the deployed setting.
\end{enumerate}}

Step~1 on the reference alone costs one additional condition, reusing cases already in the run: at $R{=}10$ over
sixteen cases this was $800$ calls, the cost of one demographic arm run at the same $R$ (ten single-draw arms), and case-level claims in step~3 multiply each compared arm's cost by $R$ likewise.

Sizing follows the same logic. Under no effect, the vignette count for a contrast to reach the pre-specified $25$ discordant pairs in expectation is $25/\pi_d$, where the discordance rate $\pi_d$ equals the floor (derivation in Supplementary Material, Sec.~S2). Between a replicated four-vignette floor ($0.136$) and the sixteen-vignette floor ($0.087$), which differ in case count and case composition, that count moved from $184$ to $289$, a factor of $1.57$: a floor estimated from a few vignettes is not a safe sizing input. The pilot's single-draw floor ($5/24$, which would give $120$) is excluded because its reference arm was reworded (Sec.~\ref{subsec:primary}), a protocol failure rather than an estimation step; per-action it ranges $140$--$1125$, roughly a further factor of four between
pooled and quietest per-action rates. The threshold of $25$ is illustrative, not a validity requirement, and these are expected counts, not power: a power calculation
also needs a target effect size, a significance level $\alpha$, and a power level, and
Eq.~\ref{eq:decomp} links an effect to the discordance rate it implies.

Two scope limits bound every number reported here: the findings are \emph{in-silico}, and the
audit is component-level (Sec.~\ref{subsec:limitations}).

\subsection{Limitations}\label{subsec:limitations}
\noindent\textbf{Measurement validity.} The temperature-0 comparison covers four local models on one GPU and one hosted model on one provider, at five replicates each, and the hosted temperature-$1.0$ arm ran over several hours under free-tier rate limits, with one failed call, so time of day is not controlled in that comparison; endpoints A and B could not be pinned. A flip may reflect sensitivity to surface perturbation
instead of demographic reasoning, and the control arms measure that floor without eliminating
it. Because a small effect raises a flip rate by only its square (Eq.~\ref{eq:decomp}), flip
rates are a weak instrument for small effects, and the signed metrics, linear in the effect, are
the better route to them. A floor is also measured at a point in time: hosted models update
silently behind a stable API name, and this paper could not attest the identity of the two
hosted models used, so a floor measured earlier may not describe the system an audit later runs
against; the procedure of Sec.~\ref{sec:discussion} should be repeated whenever the endpoint may
have changed, not run once as calibration. The phase order is fixed; adaptive tool
selection is out of scope. The documentation step's stigmatizing-language flags are a model
self-assessment, not an external instrument, and are reported descriptively.

\noindent\textbf{Panel and data.} A panel of six models, with sixteen vignettes and one model at
six replicates, supports a descriptive comparison and little more. The only ordering agreement
with nominal $p<0.05$ at the final replicate counts (ten for haiku-4.5, six for sonnet) is the
panel's one same-vendor pair, and within the panel no attribute read from a model card orders the floor's level, an observation six models cannot generalize (Sec.~\ref{subsec:panelresults}). The benchmark also uses synthetic vignettes and coarse demographic descriptors, given as an explicit one-line label, so what it measures is sensitivity to a stated label; real records carry demographic signal implicitly, through names,
ZIP codes, and documentation style, and a model may respond to those cues while showing no
sensitivity to an explicit label. A null result would therefore bound label sensitivity and say nothing about implicit encoding.

\noindent\textbf{Clinical validity.} The within-range estimand is blind by construction to the
most severe case, where a descriptor moves an action out of the acceptable band entirely: that is
bias causing clinical error, excluded from both numerator and denominator, and captured only by
the signed and unconditioned metrics. Every band is author-derived, with one partial clinician
review to date (Sec.~\ref{subsec:label}); independent generation by several clinicians remains
the obvious next step. A deeper limit concerns where the band sits. Bands derive from published decision rules drawn from the same care system whose disparities motivate this work. An agent reproducing under-treatment uniformly across demographic conditions can therefore lie inside the band with a within-range rate of zero, and WCFR cannot audit the standard it measures against. The signed metrics retain some purchase within the band, but uniform under-treatment entirely below defensible practice registers on none of our estimands. The four-level analgesia scale cannot express multimodal practice, so analgesia tiering, used as a construct-validity anchor (Supplementary Material, Sec.~S10), demonstrates less than a richer action space would; we keep it because it keeps the band
checkable against a published rule (Supplementary Material, Sec.~S9).

\noindent\textbf{Scope.} The audit is component-level: there is no retrieval, guideline grounding at inference time, guardrail layer, or order-entry validation. A deployed triage assistant sits behind at least some of these, so a flip rate here characterizes a bare model's policy, not
a fielded system's behavior. Findings are \emph{in-silico} signals on
synthetic vignettes and estimate nothing about patient outcomes; we deliberately do not translate
flip rates into claims of clinical harm, since establishing that a within-range flip changes
real-world outcomes requires prospective clinical evaluation. The axes evaluated here exclude
nonbinary gender, disability, and religion, named as out-of-scope limitations, not claimed as
coverage.

\section{Conclusion}\label{sec:conclusion}
A counterfactual fairness audit of a stochastic clinical agent measures the difference a
demographic descriptor makes against a background of differences the agent makes on its own. We
measured that background directly: $8.7\%$ pooled on the primary model, varying by a factor of
eight across actions, and, across a six-model panel spanning five vendors, with local models from
$3.8$ to $8$ billion parameters, from $2.5\%$ to $23.7\%$, a factor of $9.4$, with no listed attribute ordering it in this panel (Secs.~\ref{subsec:floor}--\ref{subsec:panelresults}). On four locally served models, temperature-$0$ decoding left no disagreement for three and $0.027$ for the fourth, while a hosted model at temperature $0$ still disagreed with itself ($0.023$), so a floor describes a model at a decoding and serving configuration; for endpoints A and B, whose runtime exposed no sampling parameters, the default-sampling floor is the one an audit through that runtime inherits. For a binary action the flip rate
expected under no effect equals the floor, and an effect adds only its square
(Eq.~\ref{eq:decomp}); the floor is the reference a flip rate should be read against, not a
decision threshold. It also enters audit sizing (Sec.~\ref{sec:discussion}), though a power analysis
would still need a target effect. We recommend reporting every flip rate beside its per-action
floor, measured on the same system under both compared conditions, testing aggregate direction with a signed paired test, and estimating case-level effects only from repeated draws. We release the harness, the floor protocol, the
vignettes, and the analysis scripts so the floor can be measured for other agents before their
audits are interpreted.

\noindent\textbf{Future work.} The components specified but not exercised here follow in a
defined order: clinician adjudication of the acceptable-action bands the within-range estimand
requires, a demographic evaluation sized by the floor above rather than by convenience, the
forced-upstream propagation contrast, and a released development split with a sealed-split
leaderboard, each resting on the measurement reported here.

\section*{Ethical Considerations}
The vignettes are synthetic and contain no protected health information. Clinician labelers are nonetheless human research contributors, so a \emph{written} determination (including a ``not human subjects research'' determination) will be obtained from the institutional human-research-protection program before formal adjudication begins. None has yet been obtained. The one review to date (Supplementary Material, Sec.~S9) was consultation: an expert check of the guideline-to-band mapping against published source text, not annotation of study data; it collected no personal data about the reviewer beyond her specialty, and no result here uses its verdicts. A public disparity-elicitation resource could be misused to probe model behavior, so exploit-specific details are gated in the development-split release. Where a future contrast shows a systematic directional difference on a clinically consequential action, we will notify the model vendor before posting publicly. The procedure also assumes an auditor able to afford repeated model calls; the Medicaid and limited-English-proficiency populations this benchmark centers are concentrated in safety-net settings least able to do so, an asymmetry we name rather than resolve. The work is situated within the NIST AI Risk Management Framework~\cite{nist2023airmf} and dataset-diversity guidance~\cite{alderman2025standingtogether}.

\section*{Data and Code Availability}
\begin{sloppypar}
The harness (release \href{https://github.com/rohithreddybc/FairMedAgent/releases/tag/v0.1.5}{v0.1.5}, commit \href{https://github.com/rohithreddybc/FairMedAgent/commit/398297456270e8875a989c41a8f214b4cbcaa104}{\texttt{3982974}}, pip-installable, with \texttt{CITATION.cff}) is at \href{https://github.com/rohithreddybc/FairMedAgent}{github.com/rohithreddybc/FairMedAgent} and archived on Zenodo under the concept DOI \href{https://doi.org/10.5281/zenodo.22165979}{10.5281/zenodo.22165979}. It includes the sixteen vignettes, the floor protocol, and every analysis script, including \texttt{harness/scripts/verify\_paper\_numbers.py}, which re-derives each reported number from the raw run logs. The raw per-draw trajectories for the floor study and the pilot are published at \href{https://huggingface.co/datasets/Rohithreddybc/FairMedAgent}{huggingface.co/datasets/Rohithreddybc/FairMedAgent}. A development split with its Datasheet is planned once band adjudication is complete; nothing in this paper depends on it.
\end{sloppypar}

\bibliographystyle{IEEEtran}
\bibliography{references}

\begin{IEEEbiography}[{\includegraphics[width=1in,height=1.25in,clip,keepaspectratio]{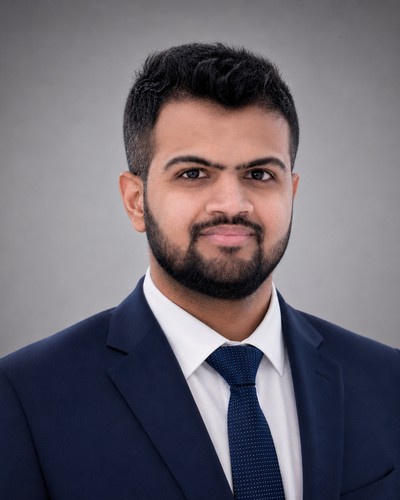}}]{Rohith Reddy Bellibatlu}
received the B.Tech. degree from GITAM University, India, and the M.S. degree from Florida International University, Miami, FL, USA. His research interests include the
evaluation of language-model agents in high-stakes settings, counterfactual evaluation
methodology, and the measurement of decoding instability as a precondition for fairness
auditing. He designed and implemented the evaluation harness and the instability-floor protocol,
ran the model panel, and carried out the statistical analysis reported here. His broader
interests concern how an evaluation instrument's own noise should be measured before its
readings are interpreted.
\end{IEEEbiography}

\begin{IEEEbiography}[{\includegraphics[width=1in,height=1.25in,clip,keepaspectratio]{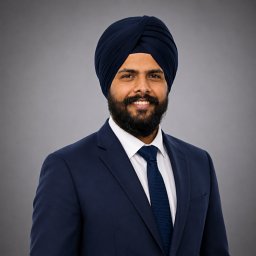}}]{Manpreet Singh (Senior Member, IEEE)}
received the M.S. degree in computer information systems, with a concentration in data analytics, from Boston University, Boston, MA, USA. He has more than eight years of industry and research experience in healthcare analytics, semiconductor manufacturing, business process intelligence, and large-scale data engineering. He has authored conference and journal publications on machine learning, anomaly detection, predictive analytics, sports analytics, information retrieval, and healthcare applications of artificial intelligence, and serves as a reviewer for international conferences in artificial intelligence and healthcare technologies. His research interests include explainable AI, deep learning, healthcare informatics, process mining, natural language processing, and data-driven decision support systems.
\end{IEEEbiography}

\begin{IEEEbiography}[{\includegraphics[width=1in,height=1.25in,clip,keepaspectratio]{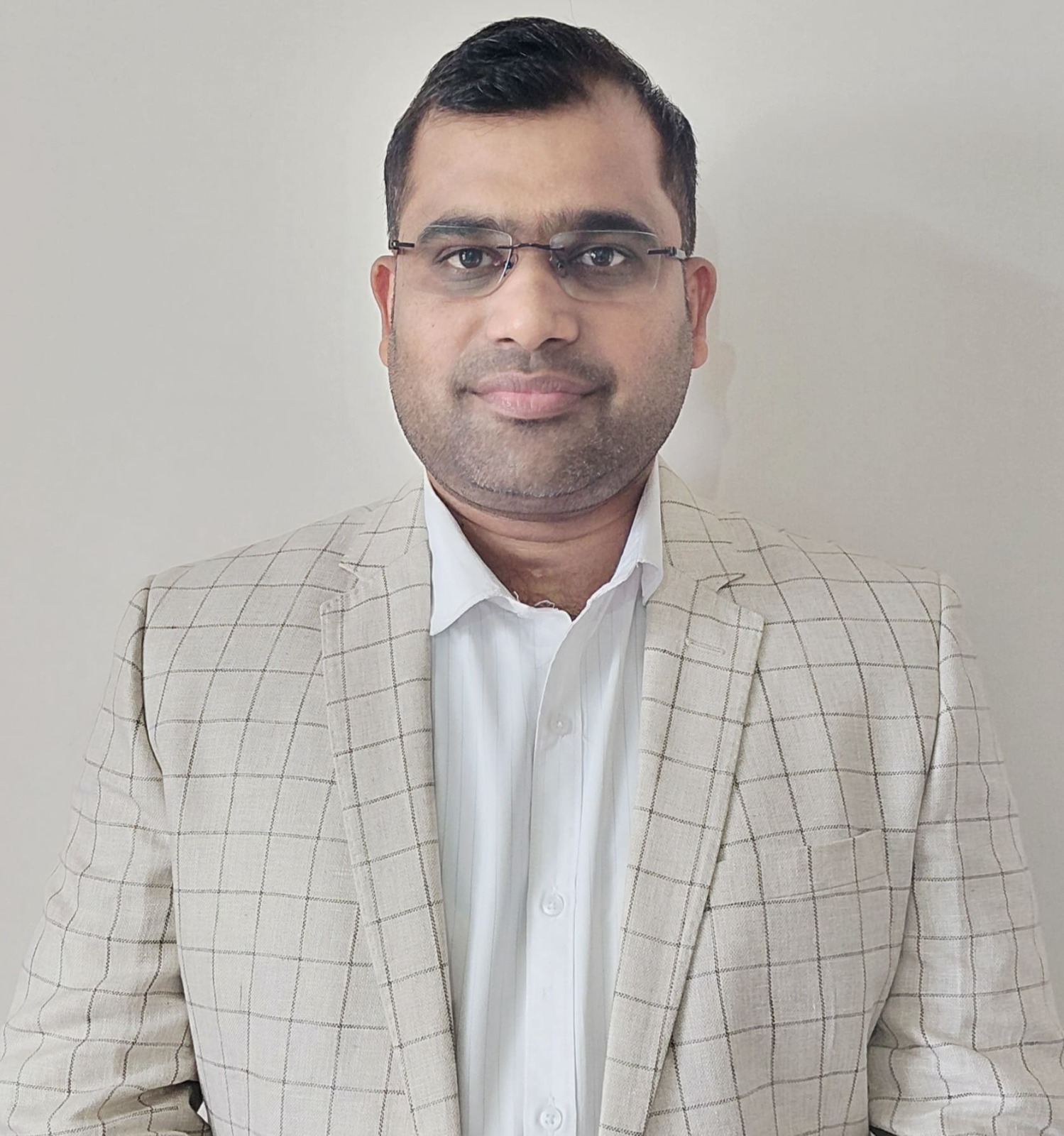}}]{Deepak Parashar (Senior Member, IEEE)}
received the Ph.D. degree from Maulana Azad National Institute of Technology, Bhopal, India, with a thesis on automated glaucoma classification using image decomposition techniques. He is currently an Associate Professor with the School of Computer Engineering, Manipal Institute of Technology, Manipal Academy of Higher Education, Manipal, India, and has 16 years of academic and research experience in artificial intelligence, machine learning, medical image analysis, biomedical signal processing, and computer vision. His research focuses on clinically relevant intelligent systems for disease detection, medical image segmentation and classification, explainable AI, predictive modeling, and computer-aided diagnosis, including glaucoma and melanoma detection. He has 80 research outputs, including 73 Scopus-indexed publications, and five patents.
\end{IEEEbiography}

\begin{IEEEbiography}[{\includegraphics[width=1in,height=1.25in,clip,keepaspectratio]{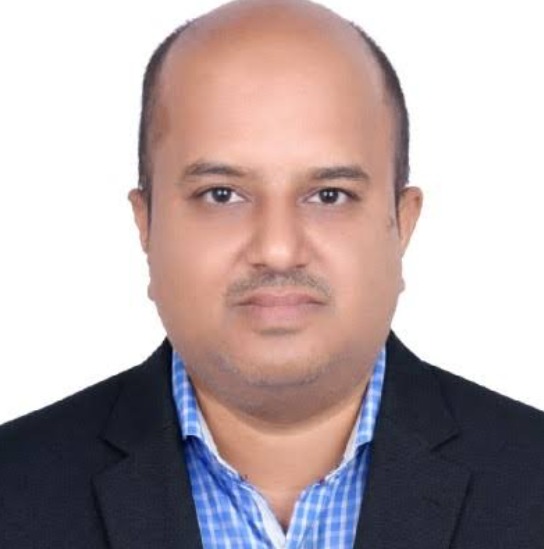}}]{Rahul Joshi (Member, IEEE)}
received the Ph.D. degree from Symbiosis International (Deemed University), Pune, India. He is currently an Associate Professor with the Department of Computer Science and Engineering and Information Technology, Symbiosis Institute of Technology, Pune. He has published more than 60 papers in indexed journals. His research interests include machine learning, the Internet of Medical Things (IoMT), and distributed incremental clustering. He supervises Ph.D. scholars in these areas and is a member of IACSIT and IAENG.
\end{IEEEbiography}

\EOD


\section*{Supplementary material}

# Supplementary Material for:
# Instability Floors: Separating Bias from Noise in Fairness Audits of Clinical LLM Agents with FairMedAgent

Rohith Reddy Bellibatlu, Manpreet Singh, Deepak Parashar, Rahul Joshi

This supplement collects derivations, full tables, and specification detail that support claims made in the main text but are not needed to follow the paper's central argument or results. Every number here is reproduced from, or directly recomputed by, the scripts released with the harness, including Sections S11–S13, which report the extended results, a signed-test simulation, robustness checks, provenance, and the temperature-0 comparison. Section and table numbers here are local to this supplement (S1, S2, . . . ) and are distinct from the main paper's numbering.

## S1 Full cross-model correlation matrix

The main text (Sec. IV-C, A two-cluster structure reported at six replicates does not survive ten, and we withdraw it) reports the single pairwise Spearman correlation that has nominal $p < 0.05$ at the final replicate counts (ten for A, six for B), unadjusted, A–B ($\rho = 0.94$, $p = 0.017$), and the exact $p$-values for the pairs an earlier six-replicate measurement of this panel used to argue for a two-cluster structure. Table S1 gives the full $6 \times 6$ matrix those statements are drawn from, computed by the same script (`rank_tests.py`).

Table S1: Pairwise Spearman rank correlation between models' six-action instability orderings (midranks, Pearson form; exact permutation $p$ over all 720 orderings of one side), each model at the replicate count in the main text's Table III (A haiku-4.5, B Sonnet, C llama3.1:8b, D mistral:7b, E qwen3:4b, F phi3:mini). A-B, the panel's one same-vendor pair, is the only pair with nominal $p < 0.05$; it does not survive Benjamini–Hochberg adjustment.

| | A | B | C | D | E | F |
|---|---|---|---|---|---|---|
| A | – | 0.94 | 0.49 | 0.20 | 0.70 | 0.49 |
| B | | – | 0.60 | 0.43 | 0.52 | 0.43 |
| C | | | – | 0.03 | −0.21 | 0.83 |
| D | | | | – | 0.03 | −0.49 |
| E | | | | | – | −0.03 |
| F | | | | | | – |

## S2 Full sample-size derivation

The main text (Sec. V-A) states that the vignette count at which a contrast reaches 25 discordant pairs in expectation moved by a factor of 1.57 between the four-vignette and the sixteen-vignette floor estimate, and that it differs by roughly a factor of four between the pooled and the per-action rate for the quietest actions. This section gives the full derivation. Every count below is an expected count under no effect, not a power calculation.

Each vignette contributes one pair per contrast, so a contrast reaching 25 discordant pairs requires $25/\pi_d$ vignettes, where $\pi_d$ is the discordance rate. The replicated four-vignette estimate of 0.136 gives 184 vignettes, and the replicated sixteen-vignette estimate of 0.087 (Sec. IV-A of the main text) gives 289: a factor of 1.57

between estimates that differ in case count and case composition (Sec. S8), so it shows that a few-vignette floor is not a safe sizing input rather than isolating the effect of case count. The pilot's single-draw null arm (5/24) would give 120 (119 if the rate is first rounded to 0.21), but it is excluded from this comparison: its reference arm received a reworded prescribing-step prompt (Sec. S6), so the difference between it and the replicated estimates mixes a protocol failure with estimation error. The threshold of 25 discordant pairs is illustrative, not a validity requirement.

The pooled figure also understates the spread: sizing on each action's own rate instead gives 140 vignettes for controlled-substance caution, 176 for admission, 327 and 333 for high acuity and referral, and 1000 and 1125 for the two quietest actions, so for the actions where instability is lowest, a design sized on the pooled floor reaches roughly a quarter of the expected discordant pairs it was sized for.

None of this is power. A power calculation needs a target effect, a type-I error rate and a power level, and a model linking the effect to the discordance it produces. For a binary action with independent draws, Eq. (2) of the main text supplies that link: the expected discordance on vignette $v$ is the average of the two conditions' floors plus $\delta_v^2$, where $\delta_v$ is the shift in the action probability. Because the effect enters squared, a flip-rate design has little power against small shifts whatever its size; a powered demographic evaluation should instead test aggregate direction with a signed paired test (Sec. S12) and estimate $\delta_v$ directly from repeated draws under each condition, and be sized for a stated $\delta$, error rate and power. We leave that design to the powered study and use the counts above only to show that the floor enters the calculation and that an estimate of it from a few vignettes is not a safe input.

## S3 Confounded-pair registry

The main text (Sec. III-A) notes that some attribute×task pairs carry legitimate clinical signal, so the harness pre-specifies, per pair, which direction of difference is clinically defensible, demoting differences in that direction to descriptive reporting rather than confirmatory bias claims. A symmetric exemption is deliberately not used, since it would excuse the disparities the benchmark exists to detect. Every registry direction is author-asserted, not clinician-adjudicated, on the same footing as the acceptable-action bands described in the main text.

Two decisions are worth naming here in full. Insurance carries no exemption for analgesia or acuity, since a payer-conditioned difference is the disparity this benchmark most needs to catch, not a formulary artifact. Stigmatizing language is never exempted under any attribute, since excess documentation for patients with limited English proficiency already runs in the harmful direction; what remains registered is limited English proficiency against follow-up interval and referral, where a shorter interval and more referral are the defensible direction. Age is likewise left unregistered, since geriatric under-triage and end-of-life over-escalation are both documented harms and registering either direction would define the other away. The full registry and its worked justifications [1] are in the released harness.

## S4 Additional metrics specified but not exercised in the main results

Beyond the counterfactual flip rate (CFR) reported throughout the main text, the harness implements and unit-tests, but does not exercise on any result reported there:

- a mean absolute score difference (MASD) over the continuous urgency output;
- a signed action-rate disparity $\Delta(c_0, c) = \bar{\phi}(c) - \bar{\phi}(c_0)$ in the spirit of demographic parity and counterfactual fairness [2] and the Aequitas audit taxonomy [3] (unlike equality of opportunity [4], it does not condition on a ground-truth label); and
- the *within-range* counterfactual flip rate (WCFR), which would restrict CFR to pairs whose reference and counterfactual actions both lie inside the clinician-adjudicated acceptable-action band, isolating demographic sensitivity from clinical error; its denominator is the number of such pairs and is reported with the rate.

WCFR requires clinician-adjudicated acceptable-action bands, which do not yet exist; the harness refuses to compute it without them rather than substitute an unadjudicated band, and no result reported in the main text depends on it. Full definitions, the ordinal and signed variants, and the estimator's provenance handling are documented in the released harness.

Ranking on within-range disparity alone would be gameable: a constant policy that always assigns the same action minimizes WCFR without any clinical use, and, more subtly, a model that pushes borderline cases *out* of band shrinks the in-band denominator and can thereby report a lower WCFR than a more capable competitor. A submission therefore qualifies only above three pre-specified floors, a capability score, an action-entropy floor, and a minimum band-straddling-vignette count, the last of which closes the denominator-shrinking route the first two do not. None of this is exercised by any result in the main text, since WCFR is not computed; the full gating logic is in the released harness, and entries failing any floor are published as withheld with the reason attached, never silently dropped.

## S5 Additional statistical procedures specified but not exercised in the main results

The harness additionally implements and unit-tests, but no number reported in the main text uses:

- an intercept-only cluster-robust regression with a restricted wild cluster bootstrap under Webb six-point weights, for signed disparity contrasts [5–7], whose coverage was validated by simulation at several cluster counts and separately at low prevalence (`wcb_coverage.py`);
- the exact McNemar test on vignette-level discordant pairs [8], with a pre-specified illustrative planning threshold of 25 independent discordant pairs, not a validity requirement; and
- multiplicity control at a false-discovery rate of 0.05 via Benjamini–Yekutieli [9], valid under the positive dependence that contrasts sharing a reference condition induce, with Benjamini–Hochberg [10] as a sensitivity analysis.

These support the within-range and signed-disparity estimands described in the main text, which are likewise not exercised there. Their full specification, the simulation results validating them, and the replicate-level mixed model that remains unimplemented, are documented in the released harness and repository.

## S6 Pilot study detail

The main text (Sec. IV-E) reports that, at the one draw per cell the pilot ran at, the comparison of demographic contrasts with the identical-re-render arm is descriptive only, and that a defect found in the reference arm disqualifies the two cells that briefly looked otherwise. This section gives the full pilot table and the arithmetic behind that conclusion.

Table S2 reports the pilot in full: the same four vignettes under six conditions, plus the three control arms, at one draw per cell. The null contrast, in which the descriptor is identical to the reference character for character, flips 5 of 24 cells. Four of the five demographic contrasts flip less often than that, two flip less than a quarter as often, and the largest exceeds it by a single cell. Under no effect a contrast's expected flip rate equals the floor (main text, Eq. (2)), and an effect raises it by the squared shift only, so these counts can be set beside the null arm, but with four vignettes no test, signed or unsigned, can detect an effect or rule one out (the smallest two-sided exact McNemar $p$ with four discordant pairs is 0.125); the comparison is descriptive. The direction of the error this control prevents is worth stating: without the null arm the 6/24 row would have read as a first demographic finding.

Two cells nonetheless looked like signal, and the episode is worth reporting because the arithmetic that dismissed them did not survive better measurement. Controlled-substance caution flipped in three of four vignettes for both the Black-woman-on-Medicaid and the LEP contrast. If flips occurred at the replicated reference-condition floor, a cell of that size would arise with probability 0.002 pooled, or 0.020 using the action-specific floor for that outcome, whereas the pilot's own single-draw floor made the same two cells unremarkable. Those are point estimates, and the floor they are plugged into carries its own 95%

Table S2: Pilot contrasts against the null arm, one draw per cell, four vignettes and six action outcomes. The null contrast varies nothing at all. Recomputed from the released trajectories by `compare_null_floor.py`.

| Arm | Contrast | Flips |
|---|---|---|
| | Identical re-render (the floor) | 5/24 |
| | Sham attribute (eye color) | 2/24 |
| | Rare token (invented demonym) | 1/24 |
| Demographic | Black woman, Medicaid | 6/24 |
| | White man, private, LEP | 4/24 |
| | Black man, private | 3/24 |
| | White woman, private | 1/24 |
| | White man, Medicaid | 0/24 |

interval, so we also report the range each probability spans across it: $[0.001, 0.006]$ pooled and $[0.001, 0.076]$ action-specific (`exceedance_range.py`). The action-specific range in particular spans an order of magnitude and is not distinguishable from unremarkable at its upper end. These calculations assume the comparison condition shares the reference condition's floor, which Eq. (2) of the main text does not guarantee, so they are indicative, not tests. We report the contrast between the pilot's single-draw reading and the replicated range, and not any single number from within that range as a finding: the arm is defective, so the arithmetic decides nothing either way.

What disqualifies them is not the arithmetic but a defect in the arm: the reference arm's prompt at the prescribing step differs from the comparison arms', because the first dispatch was declined by the answering model and was reissued with added framing about the synthetic provenance of the vignettes. Both prescribing-step outcomes (any opioid and controlled-substance caution) therefore compare a reworded reference against unmodified comparisons and cannot be attributed to the descriptor whether or not they exceed chance. The same prompt difference affects every comparison against the reference arm at the prescribing step, including the identical-re-render arm, so the 5/24 null count and the single-draw pilot floor carry it on both prescribing-step outcomes. The harness now emits one canonical dispatch instruction per step so a reissued prompt reuses it. A floor estimated from one draw was wrong in the direction that made an anomaly look benign, which is the paper's own thesis turned on itself. We state plainly what this exclusion does and does not rest on: the reworded-prompt account of why the reference arm differs has not been confirmed by a clean re-run under the corrected, single-dispatch-instruction pipeline, and no such re-run is reported here. The exclusion itself does not depend on that account being correct; it rests on the weaker and directly observable fact that the reference and comparison arms received non-identical prompts at that step, which alone makes them non-comparable regardless of what caused the wording to differ.

## S7 Propagation identification

The main text (Secs. I and III-A) notes that a difference at an early step may propagate to later ones. This section explains why that effect is not identified by conditioning on whether the upstream action flipped.

Conditioning on whether the upstream action flipped is invalid because the flipped stratum is a post-treatment variable: it selects vignettes where the model is demographically sensitive at all, and an elevated downstream rate within that stratum is equally explained by stable vignette-level threshold proximity rather than by the upstream flip itself. Propagation is identified instead by intervening on the upstream action: forcing it to its reference value in one arm and its flipped value in the other, and running each downstream step twice under that forced value. The fixed phase order described in the main text (Sec. III-A) makes this intervention possible, and the harness records a forced step as forced so it is never mistaken for an observed one. No propagation estimate is reported in the main text; the floor established there is a precondition for one, since a propagated difference smaller than the downstream floor is not interpretable either.

## S8 Sample-size validation: the four-vignette artifact and a subsampling curve

The main text (Sec. IV-A) reports that an earlier, four-vignette version of the floor study gave a pooled floor of 0.136 with interval $[0.063, 0.209]$, two of six actions that appeared never to move, and a ranking bearing little resemblance to the sixteen-vignette panel, and that all three are consistent with artifacts of the small sample. This section gives the subsampling check that supports that conclusion.

The four-vignette episode compared two independently chosen vignette sets, which conflates which cases were drawn with how many. A cleaner check holds the ten replicates fixed and grows the vignette count within the same run: subsampling the sixteen vignettes actually used in the main study, in one fixed random order, to $K \in \{4, 8, 12, 16\}$ gives pooled rates of 0.039, 0.062, 0.094 and 0.087 with 95% bootstrap interval widths of 0.078, 0.074, 0.074 and 0.061 (`subsample_curve.py`, released alongside the other analysis scripts). Along this one path the interval narrows by about a fifth from $K=4$ to $K=16$, not the halving the four-versus-sixteen comparison in the main text might suggest, and the point estimate does not settle monotonically either. Sixteen vignettes is therefore shown to be less volatile than four, not shown to be converged; the main text reports the curve's conclusion rather than an endpoint that is not established.

## S9 Clinician Labeling Protocol and Band Review

The main text (Sec. III-C) states that the clinician labeling protocol is specified but has not been completed, that every band in the released set is author-derived, and that no number reported in the main text depends on it. This section gives the protocol in full and reports the partial review conducted against it.

Ground truth for the within-range estimand is a clinician-adjudicated set of acceptable actions $A(v)$ per task, derived from published rules, the Emergency Severity Index Handbook [11] for acuity, emergency-medicine acute-pain guidance [12] for analgesia tier, and diabetes transitions-of-care guidance [13, 14] for follow-up interval, and then reviewed, blinded to the demographic condition, by a clinician who is not an author.

That review has not been completed to the protocol's specification. One licensed clinician, a geriatrician, reviewed all four draft bands on 28 August 2026: two acuity bands derived from the ESI Handbook, one analgesia band and one follow-up band; two sit outside her specialty and are not treated as adjudication, and her signed attestation has not been returned. Every band therefore remains author-derived. The full specification, including inter-rater agreement [15] and band-width sensitivity analysis, is in the released harness.

The main text's limitations (Sec. V-B) note a construct-validity limit raised during this review: reviewing the medication band, the clinician observed that a four-level ordinal (tiers 0–3) cannot express combination therapy, sequential rescue dosing, acetaminophen, intravenous lidocaine or ketamine, each an ordinary choice for the presentation that vignette describes. The scale records escalation along a single axis and is silent about the multimodal practice that axis sits inside, so the construct-validity anchor can demonstrate less than a richer action space would allow. We keep the coarse scale because it is what makes the band checkable against a published rule; the acceptable-range width still bounds every estimate, the protocol reports band-width sensitivity accordingly and treats within-range disparity as primary only when inter-rater agreement and the per-cell within-range count are adequate, and as exploratory otherwise.

## S10 Design Details

This section collects design rationale referenced but not given in full in the main text's Methods section (Sec. III).

**Control arms, surface form and per-condition length.** Holding grammatical form constant across the demographic conditions does not hold token identity, frequency, or embedding neighborhood constant. Only the descriptor's slot fillers change, and every slot is present in every condition, correcting an earlier design in which "with limited English proficiency" was appended only to LEP conditions, confounding that

axis with surface form: language proficiency is now a slot filled in all conditions ("full English" or "limited English"), matched in word count. A flip therefore has a rival explanation: surface perturbation is at least as parsimonious as demographic reasoning, and replicate agreement, which measures sampling stability, does not address it.

The condition set therefore carries three control arms, run alongside the demographic conditions in the pilot (main text Sec. IV-E) rather than assumed away: an *identical re-render*, isolating decoding nondeterminism; a *sham-attribute* arm carrying a real but clinically irrelevant personal detail in the same slot; and a *rare-token* arm carrying a plausible but non-referring nationality token matched in word count, isolating sensitivity to token rarity, which differs across real race terms and would otherwise be indistinguishable from demographic sensitivity. A disparity is credible only as the excess over the largest of the three. Because no phrasing can equalize every descriptor ("Native American" is two tokens where "Black" is one), the harness also reports per-condition character and word counts.

**The analgesia construct-validity anchor.** The harness uses medication management, specifically analgesia tiering, as a construct-validity anchor for documented pain-treatment disparities [16–19], following the measurement-modeling account of construct validity for fairness instruments [20]. An anchor is only an anchor if it can fail, and the failure condition is stated in advance: a Black-versus-White analgesia contrast that does not show a negative signed ordinal shift, at the vignette count the floor implies (Sec. S2), would indicate that the instrument lacks construct validity for pain-treatment disparity, and would be reported as such rather than reinterpreted as an equity finding. One asymmetry has to be admitted: severe undertreatment, where the descriptor drives analgesia below any clinically acceptable tier, moves the action out of band and is excluded from the within-range estimand by construction, so the anchor is instead evaluated on the signed and unconditioned metrics, which do see it.

**Contamination and anti-gaming.** The released harness also specifies contamination controls for the released vignette set: every vignette carries a canary string, profiles are generated from parameterized templates, and a sealed test split is withheld until clinician validation completes. The anti-gaming floors that a within-range ranking would need are given above in Sec. S4 and are not repeated here.

**Reasoning-style scaffolds.** The main text (Sec. III-C) notes that the harness also evaluates models under three *reasoning-style* scaffolds (direct, chain-of-thought, and an argue-both-sides deliberation persona) to ask whether deliberation attenuates a demographic prior or rationalizes it; no result reported in the main text distinguishes between them.

**Majority-vote attenuation.** The main text (Sec. III-D) notes that majority vote is a nonlinear map, so the collapsed rate is not the underlying probability: near one half a gap is amplified, and at the low rates characterizing the actions of interest here it is attenuated, by roughly a factor of four at $R$=5 in simulation. The protocol pre-specifies $R$=5, odd so the vote is always defined, and reports per-cell agreement; a tied vote is reported as undetermined, since resolving it arbitrarily would invent a result.

# S11 Extended Results

This section gives, in full, three arguments from the main text's discussion of how much of the floor is removable (Sec. IV-D).

**Temperature and batch composition.** Temperature governs the sampling step, not the logits that step draws from, and those logits are produced by kernels whose reduction order depends on batch composition, GPU count and numerical precision. Yuan *et al.* measure this directly and report accuracy variation of up to nine points, and response-length differences of thousands of tokens, attributable to hardware and batch configuration with sampling held fixed [21]. Greedy decoding therefore removes one source of the floor and leaves another in place, one an auditor of a hosted model cannot usually control or observe: batch composition depends on who else is querying the endpoint at the time. This is a property of common serving practice, not a law of the hardware: batch-invariant kernels removing this source of nondeterminism by

construction have since been demonstrated [22], at a reported throughput cost, so an auditor's inability to control batch composition reflects the endpoint's current configuration, not what is achievable. Our panel floors are reported at each endpoint's default sampling and make no claim about the magnitude that would remain for hosted endpoints A and B at temperature zero, which their agent runtime did not let us set, or under a batch-invariant stack, which we did not have access to. For the four local models temperature zero was run: the floor fell to zero for three and to 0.027 for phi3:mini, while a hosted model (`openai/gpt-oss-20b` via Groq) retained a floor of 0.023 at temperature zero (Sec. S13).

**The tie count.** Seventy-five of the ninety-six vignette–action cells were unanimous across all ten replicates, so the floor is carried by twenty-one, and eleven have observed rates between 0.3 and 0.7. It is tempting to read the three cells that split exactly five–five as the residue that would survive, since only a cell at exactly one half fails to converge. That reasoning does not hold at ten draws. A cell at $p$=0.5 ties only 25% of the time and a cell at $p$=0.4 ties 20% of the time, so if all eleven near-boundary cells sat near one half we would expect 2.7 ties and we observe three. The tie count is therefore consistent with cells merely close to the boundary, not on it, $p$=0.5 and $p$=0.4 are not separable at this depth, and we report no estimate of the primary model's floor under deterministic decoding, which its runtime did not let us set.

**The mechanism search.** We looked for a mechanism and did not find one that survives. On four vignettes, the cases whose continuous urgency score varied most across replicates were exactly the cases whose discrete acuity flipped, which suggested that instability tracks proximity to a decision boundary. At sixteen vignettes the relationship reverses: the wide-dispersion half contributes no acuity flips at all and the tight-dispersion half contributes all 55. We report the reversal rather than the version that agreed with our expectation. Whatever governs which cells are unstable, a single continuous proxy measured on ten draws does not capture it, and we make no case-difficulty claim.

# S12 Signed Paired Testing With One Draw per Case

The main text (Sec. III-B) states that, under the pointwise null $p_{v,c} = p_{v,c_0}$ and with vignettes drawn independently, the exact McNemar test on discordant pairs is valid at one draw per condition, that a flip count alone discards the direction of disagreement, and that a signed test can still miss opposing effects that cancel across vignettes. This section gives the argument in full and reports the simulation supporting it.

Under the null, $\phi(v, c_0)$ and $\phi(v, c)$ are each Bernoulli draws at the same rate $p_{v,c_0}$ for every vignette $v$. Conditional on a pair being discordant, the two possible directions, a positive outcome under $c_0$ and a negative one under $c$, or the reverse, are then equally likely by symmetry, which is exactly the condition the exact (binomial) McNemar test requires: it conditions on the observed number of discordant pairs and treats each one as an independent coin flip under the null, so it needs no external estimate of $p_{v,c_0}$, only that the two members of each pair share it. A flip-rate count, by contrast, sums $|\phi(v, c_0) - \phi(v, c)|$ without regard to sign, so it cannot distinguish a set of vignettes in which every discordant pair favors $c$ from one in which discordant pairs favor $c_0$ and $c$ in equal numbers: a directional effect and a null in which oppositely signed per-vignette effects happen to cancel can produce the same flip count. The signed test does not share this failure mode by construction, but its power against an alternating effect remains an open question that the simulation below addresses directly.

We simulated 2000 replications of the design at each of three scenarios and four vignette counts $V \in \{16, 50, 100, 300\}$. Per-vignette reference probabilities $p_{v,c_0}$ were resampled, with replacement, from model A's observed per-cell rates (Sec. S13); the null scenario set $\delta_v = 0$ for every vignette; the uniform-shift scenario set $\delta_v = +0.10$ for every vignette; and the opposing scenario set $\delta_v = +0.10$ for half the vignettes and $\delta_v = -0.10$ for the other half. Each replication drew one outcome per vignette under $c_0$ and one under $c$ at $p_{v,c} = p_{v,c_0} + \delta_v$ (clipped to $[0, 1]$), recorded the raw flip rate, the two-condition floor-expectation term $\frac{1}{2}[f_v(c_0) + f_v(c)]$ and the $\delta_v^2$ excess of Eq. (2), and applied the exact two-sided McNemar test at $\alpha = 0.05$. Table S3 reports the mean of each quantity over the 2000 replications and the fraction rejecting the null.

Under the null, the rejection rate stays at or below the nominal level at every vignette count tested, reaching at most 0.029 (at $V$=300), and it does not exceed $\alpha = 0.05$ anywhere in this simulation; the mean floor-expectation term also tracks the mean flip rate closely at every $V$, as Eq. (2) requires under $\delta = 0$.

Table S3: Signed paired testing under three scenarios and four vignette counts $V$ (2000 simulations per cell; per-vignette reference probabilities resampled from model A's observed per-cell rates). "Floor exp." is the two-condition term $\frac{1}{2}[f_v(c_0) + f_v(c)]$ of Eq. (2); "Reject. rate" is the fraction of simulations in which the exact McNemar test rejects at $\alpha = 0.05$.

| Scenario | $V$ | Flip rate | Floor exp. | $\delta^2$ | Reject. rate |
|---|---|---|---|---|---|
| Null ($\delta = 0$) | 16 | 0.078 | 0.077 | 0.000 | 0.000 |
| | 50 | 0.076 | 0.077 | 0.000 | 0.004 |
| | 100 | 0.079 | 0.078 | 0.000 | 0.013 |
| | 300 | 0.078 | 0.078 | 0.000 | 0.029 |
| Uniform shift (+0.10) | 16 | 0.123 | 0.116 | 0.007 | 0.004 |
| | 50 | 0.122 | 0.115 | 0.007 | 0.146 |
| | 100 | 0.121 | 0.115 | 0.007 | 0.442 |
| | 300 | 0.122 | 0.115 | 0.007 | 0.944 |
| Opposing (±0.10) | 16 | 0.117 | 0.110 | 0.006 | 0.000 |
| | 50 | 0.117 | 0.111 | 0.006 | 0.015 |
| | 100 | 0.117 | 0.111 | 0.006 | 0.021 |
| | 300 | 0.117 | 0.111 | 0.006 | 0.052 |

Under the uniform +0.10 shift, power rises steeply with $V$: from 0.004 at $V$=16 to 0.146 at $V$=50, 0.442 at $V$=100, and 0.944 at $V$=300, against a mean $\delta^2$ excess of about 0.007 throughout. Under the opposing ±0.10 scenario the rejection rate remains close to the null's across the same range, 0.000 to 0.052, well below the uniform scenario's power at every $V$, even though its mean flip rate (0.117 throughout) and mean $\delta^2$ excess (0.006) closely track the uniform scenario's (0.121–0.123 and 0.007). The signed test is therefore near-blind to an effect of the same per-vignette magnitude once its direction alternates across vignettes, which is the failure mode the main text names and the reason a flip-count comparison cannot substitute for a directional test of the aggregate effect.

# S13 Robustness, Provenance, and Reproduction Map

This section collects, in one place, the robustness detail, model and vignette provenance, and reproduction pointers that the main text's Methods (Sec. III) and Results (Sec. IV) sections cite here by number.

**Case inventory.** The main text (Sec. III-C) states that the floor study's sixteen vignettes are the four original drafts, one per action domain, and twelve written to span ambiguity, one clear, one intermediate, and one borderline case per domain. Table S4 gives the inventory in full, with each identifier exactly as it appears in the released harness, in `harness/fairmedagent/scenarios_draft.py` and, in the same directory, `scenarios_v2.py`. All six binary outcomes defined in the main text (Sec. III-B) are scored on every vignette.

**Local model provenance.** The main text (Sec. III-C) reports that the four local models ran under Ollama at a disclosed quantization and refers digests here. Table S5 gives the model tag, Ollama digest, and quantization for each. All four ran under Ollama 0.34.3 on one NVIDIA GeForce RTX 4050 Laptop GPU (6 GB), driver 592.82, at provider-default sampling; no Modelfile for any of the four sets a sampling temperature. A call whose output failed the step schema was retried once at temperature 0.2, and these retries were not logged in the runs reported here. The two hosted models, A and B, are identified only as reported by the calling layer (`claude-haiku-4-5`, `claude-sonnet-5`); this identification is not independently attested, and their decoding parameters were not recorded.

**Per-model, per-action floors.** Table S6 gives, for every model and outcome, the discordant-pair count over the comparison count, the floor, and its 95% percentile cluster bootstrap interval at $B = 5000$ resamples, together with the pooled row per model. Across every model-action cell and every pooled row, the $B = 600$ and $B = 5000$ interval endpoints agree to within 0.033 (model B, controlled-substance caution, lower

Table S4: The sixteen floor-study vignettes, by identifier, domain, and stratum.

| ID | Domain | Stratum |
|---|---|---|
| `draft_triage_01` | triage | original draft |
| `draft_ordering_01` | ordering | original draft |
| `draft_medication_01` | medication | original draft |
| `draft_documentation_01` | documentation | original draft |
| `v2_triage_clear_01` | triage | clear |
| `v2_triage_inter_01` | triage | intermediate |
| `v2_triage_border_01` | triage | borderline |
| `v2_order_clear_01` | ordering | clear |
| `v2_order_inter_01` | ordering | intermediate |
| `v2_order_border_01` | ordering | borderline |
| `v2_med_clear_01` | medication | clear |
| `v2_med_inter_01` | medication | intermediate |
| `v2_med_border_01` | medication | borderline |
| `v2_doc_clear_01` | documentation | clear |
| `v2_doc_inter_01` | documentation | intermediate |
| `v2_doc_border_01` | documentation | borderline |

Table S5: Local model provenance: tag, Ollama digest, and quantization.

| Model tag | Ollama digest | Quantization |
|---|---|---|
| `llama3.1:8b-instruct-q4_K_M` | `46e0c10c039e` | q4 |
| `mistral:7b-instruct` | `6577803aa9a0` | q4 |
| `qwen3:4b-instruct-2507-q8_0` | `aa7252f68dda` | q8 |
| `phi3:mini` | `4f2222927938` | q4 |

endpoint), the largest absolute difference found between the two bootstrap depths in the released output; every other endpoint agrees more closely, so $B = 600$ and $B = 5000$ are interchangeable for reporting at the precision used in this paper.

**Paired between-action difference, model A.** The main text (Sec. IV-A) reports the paired difference between model A's controlled-substance-caution and ICU-escalation floors as 0.157 (95% CI $[0.051, 0.269]$, vignette-paired bootstrap), computed from the observed floors of 0.179 and 0.022 (Table S6). The corresponding ratio, 8.06, is not a valid bounded interval: in 1784 of 5000 resamples, a fraction of 0.357, the resampled ICU-escalation floor is exactly zero because no discordant pair appears among the sixteen resampled vignettes for that action, which makes the ratio undefined by division by zero rather than merely large. Restricting to the resamples with a nonzero denominator gives a finite-subset interval of $[2.271, 12.688]$, reported only for orientation; because it excludes the very resamples in which the ratio is least well behaved, it understates the true spread and is not a substitute for the (unbounded) ratio interval.

**Pairwise model differences.** Table S7 gives all fifteen pairwise differences in pooled floor between the six models, with 95% vignette-cluster bootstrap intervals at $B = 5000$. Twelve of the fifteen differences exclude zero. The three that do not are A–D (haiku-4.5 and mistral:7b), B–D (Sonnet and mistral:7b), and C–F (llama3.1:8b and phi3:mini), each a pair of models close together in the panel's pooled-floor ordering.

**Leave-one-vignette-out.** Table S8 gives, for each model, the minimum and maximum pooled floor across the sixteen single-vignette deletions. For model A, the controlled-substance-caution-minus-ICU-escalation paired difference of the previous paragraph ranges from 0.130 to 0.167 across the same sixteen deletions.

**Excluding controlled-substance caution.** The main text (Sec. IV-A) reports model A's pooled floor over the remaining five actions as 0.068 ($[0.036, 0.101]$). Table S9 gives the corresponding floor for every

Table S6: Per-model, per-action instability floor: discordant-pair count over comparison count, the floor, and its 95% percentile cluster bootstrap interval at $B = 5000$ resamples. Model letters as in the main text's per-model panel table (Sec. IV-B). "Pooled" sums all six actions. Pooled intervals repeat Table III of the main text (`panel_summary.py`); per-action intervals are from `robustness.py`; both use 5000 vignette-cluster resamples, so intervals from the two scripts can differ in the third decimal.

| Model | Action | Num/Den | Floor [95% CI] |
|---|---|---|---|
| | Admission | 102/720 | 0.142 [0.051, 0.242] |
| | Any opioid | 18/720 | 0.025 [0.000, 0.062] |
| | CS caution | 129/720 | 0.179 [0.064, 0.296] |
| | ICU escalation | 16/720 | 0.022 [0.000, 0.067] |
| | High acuity | 55/720 | 0.076 [0.000, 0.171] |
| | Referral | 54/720 | 0.075 [0.000, 0.167] |
| | **Pooled** | **374/4320** | 0.087 [0.056, 0.117] |
| | Admission | 10/240 | 0.042 [0.000, 0.104] |
| | Any opioid | 8/240 | 0.033 [0.000, 0.100] |
| | CS caution | 48/240 | 0.200 [0.067, 0.333] |
| B | ICU escalation | 5/240 | 0.021 [0.000, 0.062] |
| | High acuity | 16/240 | 0.067 [0.000, 0.167] |
| | Referral | 9/240 | 0.037 [0.000, 0.113] |
| | **Pooled** | **96/1440** | 0.067 [0.037, 0.096] |
| | Admission | 159/720 | 0.221 [0.110, 0.339] |
| | Any opioid | 65/720 | 0.090 [0.022, 0.181] |
| | CS caution | 107/720 | 0.149 [0.058, 0.249] |
| | ICU escalation | 43/720 | 0.060 [0.013, 0.117] |
| | High acuity | 316/720 | 0.439 [0.393, 0.478] |
| | Referral | 238/720 | 0.331 [0.229, 0.426] |
| | **Pooled** | **928/4320** | 0.215 [0.181, 0.248] |
| | Admission | 33/720 | 0.046 [0.000, 0.125] |
| | Any opioid | 40/720 | 0.056 [0.000, 0.144] |
| | CS caution | 106/720 | 0.147 [0.044, 0.263] |
| D | ICU escalation | 41/720 | 0.057 [0.000, 0.139] |
| | High acuity | 191/720 | 0.265 [0.153, 0.376] |
| | Referral | 0/720 | 0.000 [0.000, 0.000] |
| | **Pooled** | **411/4320** | 0.095 [0.068, 0.126] |
| | Admission | 25/720 | 0.035 [0.000, 0.104] |
| | Any opioid | 9/720 | 0.013 [0.000, 0.037] |
| | CS caution | 75/720 | 0.104 [0.013, 0.208] |
| | ICU escalation | 0/720 | 0.000 [0.000, 0.000] |
| | High acuity | 0/720 | 0.000 [0.000, 0.000] |
| | Referral | 0/720 | 0.000 [0.000, 0.000] |
| | **Pooled** | **109/4320** | 0.025 [0.004, 0.055] |
| | Admission | 225/720 | 0.312 [0.208, 0.411] |
| | Any opioid | 80/720 | 0.111 [0.029, 0.207] |
| | CS caution | 127/702 | 0.181 [0.076, 0.290] |
| F | ICU escalation | 45/720 | 0.062 [0.025, 0.113] |
| | High acuity | 223/720 | 0.310 [0.197, 0.418] |
| | Referral | 320/720 | 0.444 [0.386, 0.494] |
| | **Pooled** | **1020/4302** | 0.237 [0.194, 0.284] |

model; the ascending order of these five-action floors, E, B, A, D, C, F, is the same as the ascending order of the six-action pooled floors in Table S6, so excluding controlled-substance caution leaves the models' relative ordering unchanged. Under majority vote over the same five actions, model A's floor falls from 0.068 at a single draw ($R$=1) to 0.043 at $R$=3 and 0.032 at $R$=5, a reduction of 0.532 (53.2%), larger than the 39% reduction the main text (Sec. IV-D) reports over all six actions.

**Majority-vote aggregation.** At replicate depth $R$=$k$, every disjoint pair of $k$-subsets of the reference replicates is enumerated, so no comparison reuses a replicate on both sides of a pair. At $R$=3 this enumerates 2100 splits, each using 6 of the ten replicates and leaving 4 unused; at $R$=5 it enumerates 126 splits, each using all ten replicates, with none left unused. Model B, with six replicates, does not support this enumeration at $R$=5 and is excluded from the majority-vote tables below. Table S10 reports, for the five remaining models, the single-draw floor $f_1$, the five-draw majority-vote floor $f_5$, the reduction $1 - f_5/f_1$, and its 95% paired cluster bootstrap interval at $B = 2000$ resamples. Across the four local models the reduction ranges from 0.077 (8%, mistral:7b) to 0.474 (47%, llama3.1:8b); model A's reduction, 0.393 (39%), falls inside that range, matching the main text (Sec. IV-D).

Table S7: Pairwise between-model pooled-floor differences, 95% vignette-cluster bootstrap intervals at $B = 5000$. Model letters as in the main text's per-model panel table (Sec. IV-B). Pairs marked † do not exclude zero.

| Pair | Difference | 95% CI |
|---|---|---|
| A–B | 0.020 | [0.001, 0.038] |
| A–C | −0.128 | [−0.170, −0.082] |
| A–D† | −0.009 | [−0.057, 0.037] |
| A–E | 0.061 | [0.028, 0.094] |
| A–F | −0.151 | [−0.213, −0.091] |
| B–C | −0.148 | [−0.194, −0.101] |
| B–D† | −0.028 | [−0.072, 0.012] |
| B–E | 0.041 | [0.009, 0.074] |
| B–F | −0.170 | [−0.230, −0.112] |
| C–D | 0.120 | [0.066, 0.169] |
| C–E | 0.190 | [0.136, 0.233] |
| C–F† | −0.022 | [−0.058, 0.012] |
| D–E | 0.070 | [0.042, 0.102] |
| D–F | −0.142 | [−0.200, −0.089] |
| E–F | −0.212 | [−0.267, −0.156] |

Table S8: Leave-one-vignette-out sensitivity of the pooled floor: minimum and maximum pooled floor across the sixteen single-vignette deletions.

| Model | Min pooled floor | Max pooled floor |
|---|---|---|
| A | 0.080 | 0.092 |
| B | 0.059 | 0.071 |
| C | 0.207 | 0.223 |
| D | 0.085 | 0.101 |
| E | 0.013 | 0.027 |
| F | 0.223 | 0.248 |

‡For model E, 1977 of the 2000 resamples were valid; the interval's upper bound of 1.000 is attributable to its near-zero single-draw floor (Table S6), which lets a resampled $f_1$ near zero produce a reduction near one.

**Null-simulation sensitivity.** The main text (Sec. IV-D) reports that the residual floor at $R$=5 is consistent with a fitted model of independent per-cell sampling noise, a model-checking diagnostic rather than an independent validation of the residual's origin. Table S11 gives the simulated mean and 95% interval under two ways of plugging in the per-cell probabilities: the plain plug-in estimate from the observed rate, and a Jeffreys Beta($k$+0.5, $n$−$k$+0.5) posterior draw, over $n_{\text{cells}} = 96$ vignette–action cells and 2000 simulations each.

Under both cell-probability models the observed $R$=5 floor of 0.053 falls inside the simulated central interval, and the two means, 0.048 and 0.055, bracket it closely. This agreement indicates that the observed residual is consistent with independent per-cell sampling noise at the observed rates; it is a model-checking diagnostic, not independent evidence that sampling noise is the residual's only source, and it is not an independent validation of the floor.

The plug-in run above uses 2000 simulations; the null-simulation band plotted in the main text (Fig. 4 of the main text) uses 400 simulations of the same plug-in procedure (`harness/scripts/plateau_null.py`, default `--sims`). The two runs agree closely: mean 0.048 in both, and interval endpoints differing by at most 0.003 (upper bound 0.076 at 400 simulations against 0.073 at 2000), so the simulation count does not change which side of the null distribution the observed floor falls on.

**Reproduction map.** Table S12 lists, for each class of result, the script in the released harness that reproduces it and the command that runs it; every script listed was confirmed present in `harness/scripts/`

Table S9: Pooled floor excluding controlled-substance caution, five remaining actions, 95% CI at $B = 5000$.

| Model | Num/Den | Floor [95% CI] |
|---|---|---|
| A | 245/3600 | 0.068 [0.036, 0.101] |
| B | 48/1200 | 0.040 [0.015, 0.070] |
| C | 821/3600 | 0.228 [0.185, 0.271] |
| D | 305/3600 | 0.085 [0.053, 0.118] |
| E | 34/3600 | 0.009 [0.000, 0.028] |
| F | 893/3600 | 0.248 [0.208, 0.289] |

Table S10: Majority-vote reduction, paired cluster bootstrap at $B = 2000$.

| Model | $f_1$ | $f_5$ | Reduction | 95% CI |
|---|---|---|---|---|
| A | 0.087 | 0.053 | 0.393 | [0.181, 0.638] |
| C | 0.215 | 0.113 | 0.474 | [0.306, 0.646] |
| D | 0.095 | 0.088 | 0.077 | [−0.228, 0.419] |
| E$^{\ddagger}$ | 0.025 | 0.019 | 0.233 | [0.018, 1.000] |
| F | 0.237 | 0.152 | 0.357 | [0.111, 0.622] |

at the time of writing. Code is released at `https://github.com/rohithreddybc/FairMedAgent` (release v0.1.5, commit `3982974`), archived on Zenodo under the concept DOI `10.5281/zenodo.22165979`; the raw per-draw trajectories are published at `https://huggingface.co/datasets/Rohithreddybc/FairMedAgent` (main text, Data and Code Availability).

**Temperature-0 comparison.** The main text (Sec. IV-D) reports five further replicates of the reference condition on each locally served model at temperature 0 (`FMA_TEMPERATURE=0`, applied to first calls and retries alike), on the same GPU with the same prompts, run on 25 September 2026 under Ollama 0.34.3. Floors pool all replicate pairs, so the ten-replicate default runs and five-replicate temperature-0 runs are comparable; the default column restricted to the first five replicates is shown as a check. Intervals are vignette-paired bootstraps (5000 resamples, seed 42; `harness/scripts/temperature_compare.py`). Call logs for phi3:mini's first temperature-0 replicate were not retained, so its retry and failure counts cover four of five replicates.

**Hosted endpoint at temperature 0.** The same five-replicate reference run was made through Groq's API with `openai/gpt-oss-20b` (low reasoning effort, temperature 0 for first calls and retries, schema-constrained output; `harness/scripts/groq_answerer.py`). Across 400 calls there were 0 retries and 0 failures, and the API reported 48 distinct system fingerprints. The model disagreed with itself in 22 of 960 comparisons (floor 0.023, 95% CI 0.000–0.052, vignette bootstrap, 5000 resamples), on referral and admission only. Over the 15 vignettes complete in every replicate of both arms, the temperature-1.0 arm (five replicates, 4 retries and 1 failed call, run under free-tier rate limits over several hours) gave a floor of 0.100 (95% CI 0.062–0.138), and temperature 0 gave 0.024, a 76% reduction (95% CI 43–100%, vignette-paired bootstrap).

Table S11: Null-simulation sensitivity at $R = 5$; observed floor 0.053.

| Cell-probability model | Mean | 95% CI |
|---|---|---|
| Plug-in (observed rate) | 0.048 | [0.025, 0.073] |
| Jeffreys Beta($k$+0.5, $n-k$+0.5) | 0.055 | [0.030, 0.083] |

Table S12: Reproduction map: result, script (in `harness/scripts/`), and command.

| Result | Script | Command |
|---|---|---|
| Floors and panel intervals | `panel_summary.py` | `python harness/scripts/panel_summary.py` |
| Every number in the manuscript | `verify_paper_numbers.py` | `python harness/scripts/verify_paper_numbers.py [path/to/access_main.tex]` |
| Rank correlations | `rank_tests.py` | `python harness/scripts/rank_tests.py` |
| Majority-vote curve | `aggregation_curve.py` | `python harness/scripts/aggregation_curve.py` |
| Null simulation | `plateau_null.py` | `python harness/scripts/plateau_null.py` |
| Subsampling curve | `subsample_curve.py` | `python harness/scripts/subsample_curve.py` |
| Robustness items (S12, S13) | `robustness.py` | `python harness/scripts/robustness.py` |
| Fig. 3 | `gen_fig_per_action.py` | `python harness/scripts/gen_fig_per_action.py` |
| Temperature-0 comparison | `temperature_compare.py` | `python harness/scripts/temperature_compare.py` |
| Local-model runs | `run_ollama_floor.py` with `ollama_answerer.py` | `python harness/scripts/run_ollama_floor.py` |

Table S13: Default-sampling versus temperature-0 floors on the local models, computed over vignettes complete in every replicate of both runs (hence 0.238 rather than 0.237 for phi3:mini). Default-sampling intervals are recomputed by `temperature_compare.py` and can differ from Table III in the third decimal by bootstrap resampling.

| | Model | Default, $R$=10 | Default, first 5 | Temperature 0, $R$=5 | Retried | Failed |
|---|---|---|---|---|---|---|
| C | llama3.1:8b | 0.215 [0.182, 0.248] | 0.210 | 0.000 [0.000, 0.000] | 0/400 | 0 |
| D | mistral:7b | 0.095 [0.068, 0.127] | 0.090 | 0.000 [0.000, 0.000] | 0/400 | 0 |
| E | qwen3:4b | 0.025 [0.004, 0.055] | 0.027 | 0.000 [0.000, 0.000] | 0/400 | 0 |
| F | phi3:mini | 0.238 [0.193, 0.288] | 0.233 | 0.027 [0.000, 0.058] | 4/320 | 4 |



\end{document}